\documentclass{article}

\usepackage{microtype}
\usepackage{graphicx}
\usepackage{subfigure}
\usepackage{booktabs} % for professional tables

\usepackage{hyperref}

\usepackage[accepted]{icml2023}

\usepackage{amsmath}
\usepackage{amssymb}
\usepackage{mathtools}
\usepackage{amsthm}

\usepackage[capitalize,noabbrev]{cleveref}

\usepackage[utf8]{inputenc} % allow utf-8 input
\usepackage[T1]{fontenc}    % use 8-bit T1 fonts
\usepackage{url}            % simple URL typesetting
\usepackage{amsfonts}       % blackboard math symbols
\usepackage{nicefrac}       % compact symbols for 1/2, etc.
\usepackage{xcolor}         % colors
\usepackage{color, colortbl}
\definecolor{greyC}{RGB}{180,180,180}
\definecolor{greyL}{RGB}{235,235,235}
\definecolor{shadecolor}{rgb}{0.92,0.92,0.92}
\definecolor{color1}{RGB}{255,190,122}
\definecolor{color2}{RGB}{142,207,201}
\definecolor{color3}{RGB}{190,184,220}
\definecolor{color4}{RGB}{130,176,210}
\usepackage{multicol}
\usepackage{multirow}
\usepackage{makecell}
\usepackage{wrapfig}
\usepackage{soul}
\usepackage{threeparttable}
\usepackage{dsfont}
\usepackage{enumerate}
\usepackage{amsmath,amsthm,amssymb}
\usepackage{natbib}
\usepackage{authblk}
\usepackage{wrapfig}
\usepackage{framed}
\usepackage{ulem}
\theoremstyle{plain}

\theoremstyle{definition}

\theoremstyle{remark}

\usepackage[textsize=tiny]{todonotes}

\icmltitlerunning{Towards Omni-generalizable Neural Methods for Vehicle Routing Problems}

\begin{document}

\twocolumn[
\icmltitle{Towards Omni-generalizable Neural Methods for Vehicle Routing Problems}

% It is OKAY to include author information, even for blind
% submissions: the style file will automatically remove it for you
% unless you've provided the [accepted] option to the icml2023
% package.

% List of affiliations: The first argument should be a (short)
% identifier you will use later to specify author affiliations
% Academic affiliations should list Department, University, City, Region, Country
% Industry affiliations should list Company, City, Region, Country

% You can specify symbols, otherwise they are numbered in order.
% Ideally, you should not use this facility. Affiliations will be numbered
% in order of appearance and this is the preferred way.
\icmlsetsymbol{equal}{*}

\begin{icmlauthorlist}
\icmlauthor{Jianan Zhou}{ntu}
\icmlauthor{Yaoxin Wu}{tue}
\icmlauthor{Wen Song}{shan}
\icmlauthor{Zhiguang Cao}{smu}
\icmlauthor{Jie Zhang}{ntu}
%\icmlauthor{}{sch}
% \icmlauthor{Firstname8 Lastname8}{sch}
% \icmlauthor{Firstname8 Lastname8}{yyy,comp}
%\icmlauthor{}{sch}
%\icmlauthor{}{sch}
\end{icmlauthorlist}

\icmlaffiliation{ntu}{School of Computer Science and Engineering, Nanyang Technological University, Singapore}
\icmlaffiliation{tue}{Department of Information Systems, Eindhoven University of Technology, The Netherlands}
\icmlaffiliation{shan}{Institute of Marine Science and Technology, Shandong University, China}
\icmlaffiliation{smu}{School of Computing and Information Systems, Singapore Management University, Singapore}
% \icmlaffiliation{yyy}{Department of XXX, University of YYY, Location, Country}
% \icmlaffiliation{comp}{Company Name, Location, Country}
% \icmlaffiliation{sch}{School of ZZZ, Institute of WWW, Location, Country}

\icmlcorrespondingauthor{Yaoxin Wu}{wyxacc@hotmail.com}
\icmlcorrespondingauthor{Wen Song}{wensong@email.sdu.edu.cn}

% You may provide any keywords that you
% find helpful for describing your paper; these are used to populate
% the "keywords" metadata in the PDF but will not be shown in the document
\icmlkeywords{Machine Learning, ICML}

\vskip 0.3in
]

% this must go after the closing bracket ] following \twocolumn[ ...

% This command actually creates the footnote in the first column
% listing the affiliations and the copyright notice.
% The command takes one argument, which is text to display at the start of the footnote.
% The \icmlEqualContribution command is standard text for equal contribution.
% Remove it (just {}) if you do not need this facility.

\printAffiliationsAndNotice{}  % leave blank if no need to mention equal contribution
% \printAffiliationsAndNotice{\icmlEqualContribution} % otherwise use the standard text.

\begin{abstract}
Learning heuristics for vehicle routing problems (VRPs) has gained much attention due to the less reliance on hand-crafted rules. However, existing methods are typically trained and tested on the same task with a fixed size and distribution (of nodes), and hence suffer from limited generalization performance. This paper studies a challenging yet realistic setting, which considers generalization across both size and distribution in VRPs. We propose a generic meta-learning framework, which enables effective training of an initialized model with the capability of fast adaptation to new tasks during inference. We further develop a simple yet efficient approximation method to reduce the training overhead. Extensive experiments on both synthetic and benchmark instances of the traveling salesman problem (TSP) and capacitated vehicle routing problem (CVRP) demonstrate the effectiveness of our method. The code is available at: \url{https://github.com/RoyalSkye/Omni-VRP}.
%Learning heuristics for vehicle routing problems (VRPs) has gained much attention due to its potential to explore efficient and powerful heuristics without much domain expertise. However, existing methods are typically trained and tested on the same task with a fixed size and distribution, and hence suffer from the generalization issue. 
% Neural combinatorial optimization (NCO) recently gained great attention due to the possibility of learning an efficient and powerful heuristic without much domain expertise. However, the NCO heuristics are typically trained and tested on the same task (i.e., with fixed size or distribution), and therefore suffer from the generalization issue. 
% In this paper, we study both the size and distribution generalization of neural methods in vehicle routing problems. Specifically, we consider a challenging yet realistic setting where a NCO heuristic is trained on diverse (i.e., hundreds of) tasks with various sizes and distributions. 
% We conduct extensive experiments to demonstrate the effectiveness of our method on the traveling salesman (TSP) and capacitated vehicle routing (CVRP) problems. Specifically, our method can improve the generalization of the base model on several settings and 
% across both size and distribution on the zero-shot, few-shot and active search settings.
\end{abstract}

\vspace{-4mm}
\section{Introduction}
\label{intro}
% 1. Introduce NCO
Combinatorial optimization problems (COPs) are of great importance in computer science and operation research. 
The exact methods suffer from the scalability issue due to the NP-hardness, while the heuristic ones need substantial hand-crafted rules and domain expertise for each specific problem.
% In conventional methods, heuristics are often used since it is more scalable and efficient than exact ones, which however may need substantial hand-crafted rules or domain expertise for each specific problem.
%\colorb{While the exact methods suffer from the scalability issue due to the NP-hardness, the heuristic ones may need substantial hand-crafted rules or domain expertise for each specific problem}.
% The exact methods (e.g., branch and bound) could solve it to the optimality but suffering from the scalability issue due to the exponential worst-case complexity. Although the heuristic method is an alternative for efficient approximation of (sub-)optimal solutions, it requires substantial hand-crafted rules and domain expertise in each specific problem.
Recently, the \emph{neural method}, which leverages machine learning (ML) to automatically learn or discover heuristics for a wide range of COPs, has gained much attention~\cite{bengio2021machine}.
% To tackle their flaws, the \emph{neural method} has gained much attention~\cite{bengio2021machine} recently, which leverages machine learning (ML) to automatically learn or discover heuristics for a wide range of COPs. 
By exploiting the underlying pattern among a group of COP instances, the neural method has the potential to reduce computational efforts while achieving desirable solution quality. In this paper, we focus on \emph{vehicle routing problems} (VRPs), which is a class of canonical NP-hard COPs with wide applications in transportation~\cite{garaix2010vehicle,zhou2023learning,wu2023neural} and logistics~\cite{cattaruzza2017vehicle,konstantakopoulos2022vehicle}.
The neural methods for VRPs usually employ advanced deep models (e.g., pointer network~\cite{vinyals2015pointer}, attention mechanism~\cite{vaswani2017attention} and graph neural network~\cite{bresson2017residual}) to learn heuristics for route construction or improvement with supervised learning or reinforcement learning. They have achieved competitive or even superior performance to the conventional heuristics.

% 2. However, current problem
However, most of the neural methods are trained and tested on the same task with a fixed size and distribution, and thus suffer from poor generalization. For example, the popular attention-based models~\cite{kool2018attention,kwon2020pomo} are only trained and tested on instances of fixed size (e.g., 100), with node coordinates sampled from the uniform distribution. The performance of the learned heuristic drastically decreases when it is applied to an unseen task during training (see POMO* in Table \ref{exp_1}). This generalization issue severely hinders the application of these neural methods in practice.
A simple measure for improvement is to train them on diverse data. However, covering the whole problem space is intractable (e.g., the "Catch-22" for NP-hard problems~\cite{yehuda2020s}). Thus how to effectively learn from diverse data for VRPs is still a challenging problem.

% 3. recent work about generalization, what are their limitations
% \colorb{Some work cross-size, some cross-dist. However, few work consider both size and dist. Importance of this setting: in realistic setting, ... -> valuable problem setting in VRPs. ~\cite{manchanda2022generalization} shortcomings, they use first-order to approximate second-order; do not consider scheduler so that make maximum usage of data.}
While some attempts have been made to tackle the generalization issue of neural methods for VRPs, most of them solely focus on either size~\cite{lisicki2020evaluating,bdeir2022attention,kim2022scaleconditioned} or distribution~\cite{zhang2022learning,jiang2022learning, wang2022game,geisler2022generalization,bi2022learning}. We argue that it is more realistic to simultaneously consider the generalization of size and distribution, since the real-world VRP instances (e.g., TSPLIB~\cite{reinelt1991tsplib} and CVRPLIB~\cite{uchoa2017new}) may vary in both.  
% but we empirically observe its inferior performance compared with the second-order meta-learning method (e.g., MAML~\cite{finn2017model}) in VRPs. sensitive to the scheduling of training tasks when facing with the diverse training task set. 
As a promising work, ~\citet{manchanda2022generalization} handles this challenging setting by exploiting a meta-learning technique, i.e., Reptile~\cite{nichol2018first}, which only relies on the first-order derivatives for training. 
% Theoretical and empirical evidences of its effectiveness are demonstrated on the few-shot supervised learning setting~\cite{nichol2018first}, but are still lacking on the reinforcement learning setting of which most of the neural methods in VRPs are favour. Typically, Reptile needs multiple inner-loop updates to incorporate information from higher-order derivatives (see Appendix \ref{app:analysis}) to achieve satisfactory performance (see Appendix \ref{app:meta-pomo}), and therefore it is less sample efficient.
% Moreover, they simply select training tasks by randomly sampling from the task set, which overlooks the training dynamics and cannot make the maximum usage of diverse data information.
However, it is still far from satisfaction in that, 1) Reptile is less sample efficient as it typically needs multiple inner-loop updates to incorporate information from higher-order derivatives (see Appendix \ref{app:analysis}) to achieve desirable performance (see Appendix \ref{app:meta-pomo}). Moreover, theoretical and empirical evidences of its effectiveness are only demonstrated on the few-shot supervised learning setting~\cite{nichol2018first} rather than the reinforcement learning setting, the latter of which is more favored by the neural methods for VRPs;
2) it simply selects the training task by randomly sampling from the task set in each iteration, which overlooks the training dynamics and fails to fully make use of the diverse data information.

% 4. therefore (motivation), in this paper, we consider ...; challenge, how to solve
% In this paper, we tackle the generalization across both size and distribution of VRPs for neural methods.
In this paper, we tackle the omni-generalization issue of neural VRP methods by training on diverse tasks, each of which relates to a unique size and distribution. 
According to the \emph{No Free Lunch Theorems of Machine Learning}~\cite{wolpert1997no}, it is unrealistic to train a one-size-fits-all model that could perform well on any task. Therefore, we also resort to meta-learning~\cite{vilalta2002perspective,hospedales2021meta} to learn a good initialized model for fine-tuning afterwards, while bypassing the limitations in~\citet{manchanda2022generalization}.
% Therefore, similar to recent works~\cite{manchanda2022generalization}, we follow the pre-trained then fine-tuning paradigm, which has already achieved some success in NLP~\cite{brown2020language}.
Specifically, we propose a generic meta-learning framework, which is model-agnostic and compatible with any model trained with gradient updates. 
% Empirically, it learns an initialized model with relatively good zero-shot performance, which can be efficiently adapted to new tasks during inference
% It learns a good initialization of model parameters, which can be efficiently adapted to new tasks with limited data during inference. 
% \colorb{by meta-training tasks adaptively selected via a hierarchical task scheduler.} 
It learns a good initialization of model parameters by performing meta-training on tasks, which are adaptively selected from the training task set via a hierarchical scheduler. 
The trained model is able to efficiently adapt to new tasks only using limited data during inference. 
Despite being effective, it needs the second-order derivatives to perform meta-model updates, making it computationally expensive when training on instances of large sizes. Therefore, we further develop a simple yet efficient approximation method by early stopping the usage of second-order derivatives, and only leveraging the first-order ones afterwards. 
% Empirically, we observe that our method only needs a few (e.g, one) inner-loop update(s) to achieve a strong zero-shot performance.
% However, the second-order meta-learning method can only take a few steps\footnote{We empirically find that a normal GPU can only afford one gradient step in the inner-loop optimization when meta-training POMO on large instances (e.g., with size of 200).} in the inner-loop optimization due to the computation of second-order derivatives. To solve this problem, we further improve it by stopping the backpropagation at a certain step, therefore allowing taking several extra gradient steps without the extra computational cost (i.e., GPU memory). % Moreover, our hierarchical task scheduler is able to adaptively select training task according to the current training dynamics (e.g., iteration or task hardness), enabling the meta-learning framework to find a better initialization.

% 5. Contribution summary
% \colorb{1. setting; 2. meta-learning framework and hierarchical task scheduler; 3. Exp.}
Our contributions are summarized as follows. 1) We study a challenging yet realistic setting for neural VRP methods by considering the omni-generalization across both size and distribution. 
2) We propose a generic meta-learning framework, where we leverage a second-order technique to enable effective learning of an initialized model with the capability of efficient adaptation to new tasks only using limited data during inference.
% and also present a hierarchical task scheduler to adaptively select training tasks based on the training dynamics.
3) To reduce the meta-training cost, we develop a simple yet efficient approximation method, which performs comparable to the one with full second-order derivatives.
% that replaces the second-order derivatives with the first-order ones at the early stage of meta-training.
% , and provide a comparison analysis with Reptile in our problem setting.
4) We evaluate the effectiveness of our method on the traveling salesman problem (TSP) and capacitated vehicle routing problem (CVRP) by meta-training POMO~\cite{kwon2020pomo} and L2D~\cite{li2021learning}. The experimental results demonstrate that our method could improve the omni-generalization of the base models even on the zero-shot setting. We also observe consistent superiority on the few-shot settings,
% and active search~\cite{bello2017neural,hottung2022efficient} settings, 
and on the classical benchmark datasets such as TSPLIB and CVRPLIB.

\section{Related Work}
\label{related_work}
% In this section, we first introduce the neural methods for solving VRPs. Then, we review the recent attempts to mitigate their generalization issues, followed by a brief introduction to the meta-learning and its application in NCO.

\textbf{Neural Methods for VRPs.}
% Due to the efficiency to solve VRPs without much domain expertise, neural methods have gained widespread interest recently.
%Recently, neural methods for VRPs have gained widespread interests due to the effectiveness of learning an efficient and powerful heuristic without much domain expertise.
% Due to the potential to efficiently approximate solutions to the NP-hard COPs without much domain knowledge, the use of deep learning models to learn neural combinatorial heuristics has gained attention recently~\cite{bengio2021machine}. 
Most of recent neural methods for VRPs could be divided into two categories:
1) \emph{Learning Construction Heuristics:} the solution is constructed sequentially or in a one-shot manner by the learned heuristic without iterative modifications. ~\citet{vinyals2015pointer} proposes the Pointer Network (Ptr-Net) to solve TSP with supervised learning. Subsequent works train Ptr-Net using reinforcement learning to solve TSP~\cite{bello2017neural} and CVRP~\cite{nazari2018reinforcement}. 
\citet{kool2018attention} introduces the attention model (AM) based on the Transformer architecture~\cite{vaswani2017attention} to solve a wide range of COPs including TSP and CVRP. \citet{kwon2020pomo} further proposes the policy optimization with multiple optima (POMO), which improves upon AM by exploiting solution symmetries. Besides Ptr-Net and attention-based models, graph neural networks are also exploited to solve VRPs~\cite{dai2017learning,joshi2019efficient}. Other works have also been proposed to improve upon them~\cite{ma2019combinatorial,kwon2021matrix,xin2021multi,kim2022symnco}.
2) \emph{Learning Improvement Heuristics:} an initial complete solution is iteratively refined by the learned heuristic until a termination condition is satisfied. In this line of research, the classical local search methods (e.g., 2-opt~\cite{croes1958method} or large neighborhood search (LNS)~\cite{shaw1998using}) or (part of) specialized heuristic solvers for VRPs (e.g., Lin-Kernighan-Helsgaun (LKH)~\cite{helsgaun2000effective,helsgaun2017extension}) are usually exploited ~\cite{chen2019learning,lu2020learning,hottung2020neural,d2020learning,wu2021learning,wang2021bi,ma2021learning,xin2021neurolkh,kim2021learning,hudson2022graph}. In general, the improvement heuristics could achieve better performance than the construction ones, but at the expense of much longer inference time. Besides VRPs, neural methods have also been applied to solve other COPs, such as the job shop scheduling problem (JSSP)~\cite{zhang2020learning}, maximal independent set (MIS)~\cite{dai2017learning,ahn2020learning} and boolean satisfiability (SAT)~\cite{selsam2019learning}. In this paper, we mainly focus on the neural methods for VRPs.
% For a comprehensive review, see~\cite{bengio2021machine}.

\textbf{Generalization of Neural Methods.} 
Previous works mainly train and test on instances with the same size and distribution, which results in poor generalization performance~\cite{joshi2021learning,liu2022good}. Recently, some works attempt to improve the generalization capability of neural methods for VRPs, which could be further divided into two classes: 
1) \emph{Size Generalization:} the aim is to generalize the learned heuristics to instances smaller or larger\footnote{The community cares more about the larger instances since they are usually much harder than the smaller ones.
%smaller ones could be trivially solved by conventional algorithms.
} than the training ones. 
~\citet{lisicki2020evaluating} trains a model on multiple sizes via the curriculum learning strategy~\cite{bengio2009curriculum}.
\citet{kim2022scaleconditioned} and \citet{bdeir2022attention} study the size generalization upon the attention-based models by improving their structures, such as incorporating a scale conditioned network and a sparse dynamic attention, respectively.
% ~\cite{kim2022scaleconditioned} plugs an extra network into the pre-trained model to incorporate the scale information and improve its transferability on larger-scale instances.
We also would like to mention another line of research that deals with the \emph{scalability}, which is less related to the generalization issue studied in this paper. Specifically, they mainly rely on the heat-map~\cite{joshi2019efficient,fu2021generalize,qiu2022dimes,sun2023difusco} or decomposition~\cite{li2021learning,hou2023generalize} so as to scale up to larger instances.
% ~\cite{fu2021generalize} and ~\cite{qiu2022dimes} leverage heatmap-based methods~\cite{joshi2019efficient} which parameterizes the solution space as a continuous one. 
% ~\cite{li2021learning} decomposes the whole problem into several subproblems that could be solved efficiently by a subsolver, and iteratively optimizes one at a time.  
% improves the generalization ability of the supervised learning based (i.e., heat-map based~\cite{joshi2019efficient}) model by first training a small-scale model, which is repetitively used to build the heat maps of larger instances. tackles the large-scale instances by parameterizing the solution space as a continuous one (i.e., heatmap-based methods~\cite{joshi2019efficient}).
2) \emph{Distribution Generalization:} the aim is to generalize the learned heuristics to instances sampled from various (unseen) distributions. Some works augment the training instances with diverse distributions either by jointly training an instance generator~\cite{wang2022game} or adversarial training~\cite{zhang2022learning,geisler2022generalization}. 
Other than the data perspective, \citet{jiang2022learning} and \citet{bi2022learning} exploit the group distributionally robust optimization~\cite{Sagawa2020Distributionally} and knowledge distillation~\cite{hinton2015distilling} to improve the distribution generalization, respectively.

Although some efforts have been made either for the size or distribution generalization, rare works consider both of them. \citet{manchanda2022generalization} first studies this setting using Reptile~\cite{nichol2018first}. It gradually updates the meta-model towards the task-specific model on each task such that the learned meta-model could serve as a good initialization for fine-tuning during inference. 
% with the capability of fast adaptation to new tasks during inference. 
In this paper, we also focus on this challenging yet realistic problem setting. 

\begin{figure*}[!t]
    \vskip 0.1in
    \begin{center}
    \centerline{\includegraphics[width=1.96\columnwidth]{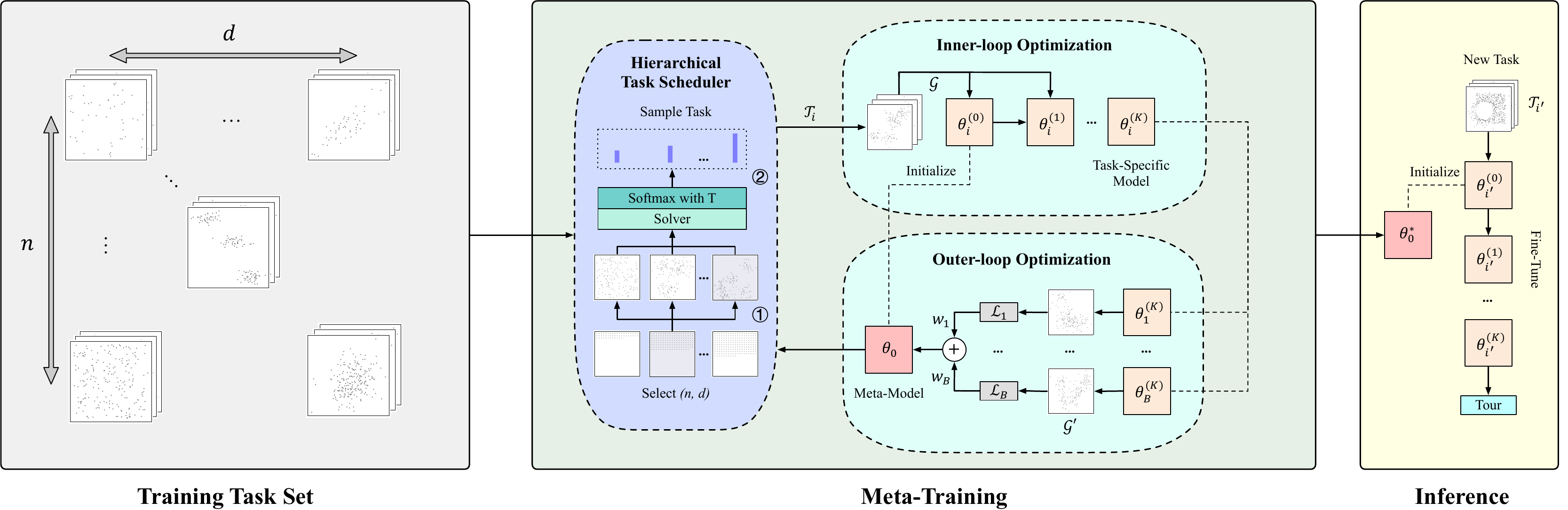}}
    \caption{The illustration of the proposed framework. The training task set consists of tasks with diverse sizes $n\sim\mathcal{N}$ and distributions $d\sim\mathcal{D}$. In each iteration, the hierarchical task scheduler adaptively selects a batch of tasks $\{\mathcal{T}_i\}_{i=1}^B$ for meta-training, which consists of a pair of inner-loop and outer-loop optimization. During inference, given a new task $\mathcal{T}_{i'}$, the trained meta-model $\theta_0^*$ is used to initialize the task-specific model $\theta_{i'}^{(0)}$, which is then adapted to $\mathcal{T}_{i'}$ by further taking $K$ gradient steps with a limited number of instances.}
    \label{overview}
    \end{center}
    \vskip -0.2in
\end{figure*}

\section{Preliminaries}
\label{prelim}
In this section, we introduce the problem statement for VRPs and the Markov Decision Process (MDP) formulation for constructing solutions to VRP instances.
% In this section, we provide preliminaries for VRPs and meta-learning. For VRPs, we introduce its problem statement and Markov Decision Process (MDP) of the solution construction process. For meta-learning, we focus on the well-known MAML proposed in ~\cite{finn2017model}.
% \subsection{VRPs and Solution Construction}
Without loss of generality, we define a VRP instance of size $n$ over a graph $\mathcal{G}=\{\mathcal{V}, \mathcal{E}\}$, where $v_i \in \mathcal{V}$ ($\mathcal{V}=\{v_i\}_{i=1}^{n}$) represents the node (e.g., customer in TSP) and $e(v_i, v_j) \in \mathcal{E}$ represents the edge between node $v_i$ and $v_j$. The solution (i.e., tour) $\tau$ is represented as a sequence of nodes in $\mathcal{V}$. In this paper, we consider the Euclidean VRPs with the cost function $c(\cdot)$ defined as the total length of the tour.
%The cost function $c(\cdot)$ computes the weighted sum of all edges in a tour, which refers to the euclidean length in this paper. 
The objective of VRPs is to find the optimal tour $\tau^*$ with the \emph{minimal} cost:
\begin{equation}
    \label{eq:obj}
    \tau^* = \arg\min_{\tau\in S} c(\tau|\mathcal{G}),
\end{equation}
where $\mathcal{S}$ is the discrete search space that contains all the feasible tours, subject to the problem-specific constraints. Specifically, a feasible tour in TSP should visit each (customer) node exactly once and return to the starting node in the end. On top of TSP, a depot node $v_0$ is introduced in CVRP, where each customer node is featured by a demand $\delta_i$, and a capacity limit $Q$ is set for each vehicle.
% a vehicle with the capacity $Q$ visits the customers in a tour. 
A tour in CVRP consists of multiple sub-tours, each of which represents a vehicle starting from the depot, visiting a subset of nodes in $\mathcal{V}$ and returning to the depot. It is feasible if each customer node is visited exactly once and the total demand in each sub-tour does not exceed the capacity limit $Q$.

Neural construction methods formulate the solving process of a VRP instance $\mathcal{G}$ as a MDP, where they typically parameterize the policy by an encoder-decoder structured neural network to learn node selection for constructing a solution. 
%Typically, it follows the learned policy $\pi_{\theta}$, which is parameterized by a deep neural network with an encoder-decoder architecture. 
The encoder in the policy network outputs a global representation of the instance, which, with the representation of the context (e.g., the partial tour in construction), captures the current state. The decoder takes the global and context representations as inputs to compute the probabilities of nodes (i.e., actions) to be visited. The node is selected sequentially until a complete tour $\tau$ is constructed. Hence, the probability of the tour is factorized via the chain rule as:
% the decoder takes as inputs together with the context one (e.g., constructed partial tour) representing the current state. Then, the decoder takes an action by selecting the next valid node to visit. The process is iterative until a complete tour $\tau$ is constructed. The probability is factorized via the chain rule as:
% At the step $t$, the state $s_t$ is characterized by global embedding of the instance and the context embedding (e.g., constructed partial solution). The action $a_t$ is to add one valid node into the current partial solution. The process is iterative in an autoregressive way until until a complete tour $\tau$ is constructed:
\begin{equation}
    \label{eq:mdp}
    p_{\theta}(\tau|\mathcal{G}) = \prod_{t=1}^{T} p_{\theta} (\pi_{\theta}(t)|\pi_{\theta}(<t),\mathcal{G}),
\end{equation}
where $\pi_{\theta}(t)$ and $\pi_{\theta}(<t)$ are the selected node and the current partial solution at time step $t$, respectively; $T$ denotes the number of total steps. The reward is defined as the negative cost of a tour, i.e., $\mathcal{R}=-c(\tau|\mathcal{G})$. 
%In order to learn such a construction heuristic, 
To train the policy network, the REINFORCE algorithm~\cite{williams1992simple} is commonly used to estimate the gradient of the expected reward $\mathcal{L}(\theta|\mathcal{G})=\mathbb{E}_{p_{\theta}(\tau|\mathcal{G})}c(\tau)$ such that:
\begin{equation}
    \label{eq:reinforce}
    \small
    \nabla_{\theta} \mathcal{L}(\theta|\mathcal{G}) = \mathbb{E}_{p_{\theta}(\tau|\mathcal{G})} [(c(\tau)-b(\mathcal{G})) \nabla_{\theta}\log p_{\theta}(\tau|\mathcal{G})],
\end{equation}
where $b(\cdot)$ is a baseline function to reduce gradient variance. 

\section{Methodology}
\label{methodology}
In this section, we introduce a generic meta-learning framework to tackle the 
omni-generalizable issue of neural methods across both size and distribution in VRPs.
% \colorb{enable learning neural heuristics for VRPs that are omni-generalizable across both size and distribution}. 
We further develop a simple yet efficient approximation method to promote the training efficiency.
%computational cost of meta-training.
% give a comparison analysis about the first-order approximation methods.
Without loss of generality, we present our method by taking the meta-training of POMO~\cite{kwon2020pomo} on TSP as an example. 

\subsection{Meta-Learning Framework}
\label{meta-framework}
We define a task $\mathcal{T} (n,d) \sim p(\mathcal{T})$ as a class of instances with the same size $n\in \mathcal{N}$ and distribution $d\in \mathcal{D}$, where $\mathcal{N}$ = $[n_{\min}, n_{\max}]$ is the range of problem sizes; $\mathcal{D}$ = $\{\mathcal{D}_j\}_{j=1}^{|\mathcal{D}|}$ is a set of distributions; $p(\mathcal{T})$ is the underlying task distribution. 
% It could be further characterized with $(\mathcal{N}, \mathcal{D})$, where $\mathcal{N}=[n_{\min}, n_{\max}]$ and $\mathcal{D}$ represent the range of sizes and distributions of all tasks, respectively. 
For notation simplicity, we use $\mathcal{T}_i$ to represent a task.
Inspired by MAML~\cite{finn2017model}, we propose a generic meta-learning framework to improve the generalization capability of neural methods for VRPs, which is model-agnostic and compatible with any model trained with gradient updates. 
%tackle the crucial generalization issue in VRPs. 
The framework is illustrated in Figure \ref{overview}, and we elaborate its key components below.
%Below, we concretely introduce key components of the framework as shown in Figure \ref{overview}. 

% \textbf{Meta-Training.} In the meta-learning framework, we aim to train a good initialization of model parameters (i.e., meta-model) $\theta_0$ that could be efficiently adapted to new tasks. Formally, we define the meta-objective as follows:
\textbf{Meta-Training.} In the meta-learning framework, we aim to train a meta-model $\theta_0$, which as a good initialized model can be efficiently adapted to new tasks during inference. Formally, we define the meta-objective as follows:
\begin{equation}
    \label{eq:meta_obj}
    \theta_0^* = \arg\min_{\theta_0} \mathbb{E}_{\mathcal{T}_i \sim p(\mathcal{T})} \mathbb{E}_{\mathcal{G} \sim \mathcal{T}_i}
    \mathcal{L}_i(\theta_i^{(K)}|\mathcal{G}),
\end{equation}
where $\theta_i^{(K)}$ is the fine-tuned model after $K$ gradient updates of $\theta_0$ on the task $\mathcal{T}_i$; 
$\mathcal{L}_i$ is the loss function on the task $\mathcal{T}_i$. In this paper, we use the same loss function (e.g., reinforcement loss) for different tasks.
To \emph{directly} optimize this objective, the meta-training procedure comprises the inner-loop and outer-loop optimization at each iteration. The meta-training pseudocode is presented in Algorithm \ref{alg:pomo-maml}.

\emph{Inner-loop optimization:} It optimizes a task-specific model iteratively, which is similar to the fine-tuning stage during inference. Specifically, given a task $\mathcal{T}_i \sim p(\mathcal{T})$, we initialize the task-specific model by the meta-model, i.e., $\theta_i^{(0)} \gets \theta_0$ (line 5), and adapt it to $\mathcal{T}_i$ by performing $K$ gradient update steps on the training instances (line 6-11).
The gradient of the loss function $\mathcal{L}_i$ at the $k_\mathrm{th}$ step is computed as follows:
%the model is updated with the average gradient of the loss function $\mathcal{L}_i$ as follows:
\begin{equation}
    \label{eq:grad}
    \small
    \nabla_{\theta_i^{(k-1)}} \mathcal{L}_i(\theta_i^{(k-1)}) \gets \frac{1}{M}\sum_{m=1}^{M}\nabla_{\theta_i^{(k-1)}} \mathcal{L}_i(\theta_i^{(k-1)}|\mathcal{G}_m),
\end{equation}
where $\nabla_{\theta_i^{(k-1)}} \mathcal{L}_i(\theta_i^{(k-1)}|\mathcal{G}_m)$ can be estimated by the REINFORCE algorithm as in Eq. (\ref{eq:reinforce}).
%the policy gradient method, e.g., Eq. (\ref{eq:reinforce}). 

\emph{Outer-loop optimization:} It optimizes the meta-model with the objective in Eq. (\ref{eq:meta_obj}). Concretely, for each task $\mathcal{T}_i$, the few-shot generalization performance of the task-specific model $\theta_i^{(K)}$ is evaluated on the validation instances. The meta-gradient (line 13) is obtained as follows: 
\begin{equation}
    \label{eq:meta_grad}
    \small
    \nabla_{\theta_0} \mathcal{L}_i(\theta_i^{(K)}) \gets \frac{1}{M}\sum_{m=1}^{M}\nabla_{\theta_i^{(K)}} \mathcal{L}_i(\theta_i^{(K)}|\mathcal{G}_m') \cdot \frac{\partial \theta_i^{(K)}}{\partial \theta_0}.
\end{equation}
After conducting the inner-loop optimization on a batch of tasks $\{\mathcal{T}_i\}_{i=1}^B$, the meta-model $\theta_0$ is then updated once (as in line 16).
% $\theta_0 \gets \theta_0- \beta \sum_{i=1}^{B} \nabla_{\theta_0}\mathcal{L}_i(\theta_i^{(K)})$. 
Intuitively, the inner-loop optimization serves as the task adaption stage, imitating the fine-tuning process during inference, while the outer-loop optimization updates the meta-model with the objective of maximizing the few-shot generalization performance of the task-specific model $\theta_i^{(K)}$. Therefore, after meta-training, we can get a good initialized model $\theta_0^*$ with the capability of efficient adaptation to new tasks only using limited data. Note that Eq. (\ref{eq:meta_grad}) needs the second-order derivatives since we expect to get the gradient direction with respect to the meta-model $\theta_0$. We will discuss and analyze its first-order approximation methods later.

\begin{algorithm}[!t]
  \small
  \caption{Meta-Training for VRPs}
  \label{alg:pomo-maml}
  {\bf Input:} distribution over tasks $p(\mathcal{T})$, number of tasks in a mini-batch $B$, number of inner-loop updates $K$, batch size $M$, step sizes of inner-loop and outer-loop optimization $\alpha, \beta$;\\
  {\bf Output:} meta-model $\theta_0^*$;
\begin{algorithmic}[1]
  \STATE Initialize meta-model $\theta_0$
%   \FOR{$e = 1$, $\dots$, $E$}
    \WHILE{not done}
    % \STATE $\{\mathcal{T}_i, w_i\}_{i=1}^B, \gets$ HierarchicalTaskScheduler$(p(\mathcal{T}), e, \theta_0)$
    \STATE $\{\mathcal{T}_i, w_i\}_{i=1}^B, \gets$ Hierarchical task scheduler
    \FOR{$i = 1$, $\dots$, $B$}
        \STATE Initialize task-specific model $\theta_i^{(0)} \gets \theta_0$
        \FOR{$k = 1$, $\dots$, $K$} 
            \STATE \COMMENT{\colorbox{shadecolor}{// \emph{Inner-loop optimization}}}
            \STATE Sample training instances $\{\mathcal{G}_m\}_{m=1}^M$ from task $\mathcal{T}_i$
            % \STATE $\nabla_{\theta_i^{(k-1)}} \mathcal{L}_i(\theta_i^{(k-1)})$ \\ $\gets \frac{1}{M}\sum_{m=1}^{M}\nabla_{\theta_i^{(k-1)}} \mathcal{L}_i(\theta_i^{(k-1)}|\mathcal{G}_m)$ using Eq. (\ref{eq:reinforce})
            \STATE Obtain $\nabla_{\theta_i^{(k-1)}} \mathcal{L}_i(\theta_i^{(k-1)})$ using Eq. (\ref{eq:grad})
            \STATE $\theta_i^{(k)} \gets \theta_i^{(k-1)} - \alpha \nabla_{\theta_i^{(k-1)}} \mathcal{L}_i(\theta_i^{(k-1)})$
        \ENDFOR
            \STATE Sample validation instances $\{\mathcal{G}_m'\}_{m=1}^M$ from task $\mathcal{T}_i$
            \STATE Obtain $\nabla_{\theta_0} \mathcal{L}_i(\theta_i^{(K)})$ using Eq. (\ref{eq:meta_grad})
            % \STATE $\nabla_{\theta_0} \mathcal{L}_i(\theta_i^{(K)}) \gets \frac{1}{M}\sum_{m=1}^{M}\nabla_{\theta_i^{(K)}} \mathcal{L}_i(\theta_i^{(K)}|\mathcal{G}_m') \cdot \frac{\partial \theta_i^{(K)}}{\partial \theta_0}$
            % \STATE $\nabla_{\theta_0}J(\theta_i) \gets \text{POMO}(\{x_m'\}_{m=1}^M,\theta_i) \cdot \frac{\partial \theta_i}{\partial \theta_0}$
            % \STATE \COMMENT{\colorbox{color2}{FOMAML:}} $\nabla_{\theta_b}J(\theta_b) \gets \text{POMO}(\{x_m'\}_{m=1}^M,\theta_b)$
    \ENDFOR
    \STATE \COMMENT{\colorbox{shadecolor}{// \emph{Outer-loop optimization}}}
    \STATE $\theta_0 \gets \theta_0- \beta \sum_{i=1}^{B} w_i \nabla_{\theta_0}\mathcal{L}_i(\theta_i^{(K)})$
    % \STATE \COMMENT{\colorbox{color2}{FOMAML:}} $\theta_0 \gets \theta_0- \beta \frac{1}{B} \sum_{b=1}^{B} \nabla_{\theta_b}J(\theta_b)$
    % \STATE \COMMENT{\colorbox{color4}{REPTILE:}} $\theta_0 \gets \theta_0 + \beta\frac{1}{B} \sum_{b=1}^B (\theta_b-\theta_0)$
  \ENDWHILE
\end{algorithmic}
\end{algorithm}

\textbf{Hierarchical Task Scheduler.} The aim of a task scheduler is to guide the task selection for the meta-training process, so as to improve the optimization performance. Most existing works  ~\cite{finn2017model,Raghu2020Rapid,flennerhag2022bootstrapped,manchanda2022generalization} randomly sample training tasks from the task set with a uniform probability, which assumes all tasks are equally important. It may overlook the training dynamics and cannot
make full use of diverse data information.
% Recently, ~\cite{yao2021meta} learns an adaptive task scheduler through a bi-level optimization, but greatly increasing the training complexity. 
In this paper, we propose a simple yet effective hierarchical task scheduler for VRPs. 

% We characterize all tasks with $(\mathcal{N}, \mathcal{D})$, where $\mathcal{N}=[n_{\min}, n_{\max}]$ and $\mathcal{D}$ represent the range of sizes and distributions in the training task set, respectively. 
During training, we first gradually increase the size of training task following a linear scheduler. 
At the $e_\mathrm{th}$ iteration of meta-training, we select tasks of the size $n_e=\lfloor n_{\min} + \min(\frac{e}{E_s},1) \cdot (n_{\max} - n_{\min}) \rfloor$, 
% \begin{equation}
%     \label{eq:n}
%     n_e = \lfloor n_{\min} + \min(\frac{e}{E_s},1) \cdot (n_{\max} - n_{\min}) \rfloor,
% \end{equation}
where $E_s$ is the working duration of the scheduler. Then, a probability distribution over the tasks in $\{\mathcal{T}(n_e, d_j)\}_{j=1}^{|\mathcal{D}|}$ is generated for training task selection, based on their hardness.
%Then, we choose the distribution $d$ based on the hardness of tasks \colorb{in} $\{\mathcal{T}(n_e, d_j)\}_{j=1}^{|\mathcal{D}|}$.
The optimality gap is an appropriate metric to measure the hardness of each task. However, it is intractable to obtain optimal solutions due to the NP-hardness. Here we use a general VRP solver LKH3~\cite{helsgaun2017extension} to efficiently\footnote{We set the maximum trials of LKH3 to 100 for efficiency.} obtain near-optimal solutions for a validation set of instances in a one-shot manner. The validation instances are fixed throughout the meta-training process, and hence we only need to run LKH3 once. To avoid overfitting, we only sample 
%a subset of $M$ validation instances 
$M$ instances from the validation set to calculate the relative gap $g_i = \frac{1}{M}\sum_{m=1}^M \frac{c(\tau_m) - c(\bar{\tau}_m)}{c(\bar{\tau}_m)}$ for each task $\mathcal{T}_i$, where $\tau_m$ and $\bar{\tau}_m$ are solutions to the $m_\mathrm{th}$ instance constructed by the current meta-model and LKH3, respectively. Accoringly, the probability of selecting each task $\mathcal{T}_i$ is defined as:
\begin{equation}
    \label{eq:weight}
    w_i = \frac{\exp (g_i/\eta)}{\sum_{j=1}^{|\mathcal{D}|} \exp (g_j/\eta)},
\end{equation}
where $\eta$ is the temperature to control the entropy of probability distribution, from which the task is sampled.
%the hierarchical task scheduler samples. 
The initial probability distribution is uniform, and is updated by Eq. (\ref{eq:weight}) periodically (e.g., every 100 iterations). After a batch of tasks $\{\mathcal{T}_i\}_{i=1}^B$ is sampled, we normalize their weights such that $\sum_{i=1}^B w_i=1$. Then, the meta-model is optimized with the weighted sum of their losses $\{w_i \mathcal{L}_i(\theta_{i}^{(K)})\}_{i=1}^B$ (line 16).
% The update of outer-loop optimization is therefore rewritten as $\theta_0 \gets \theta_0- \beta \sum_{i=1}^{B} w_i \nabla_{\theta_0}\mathcal{L}_i(\theta_i^{(K)})$ (line 16).

\textbf{Inference.} During inference, given instances sampled from a new task $\mathcal{T}_{i'}$, the trained meta-model $\theta_0^*$ could be used to approximate solutions in several ways: 
1) \emph{zero-shot:} the solution is directly constructed using the learned policy $\pi_{\theta_0^*}$ with efficient search strategies (e.g., greedy rollout);
% sampling and beam search;
2) \emph{few-shot:} similar to the inner-loop optimization as shown in line 6-11 of Algorithm \ref{alg:pomo-maml}, the task-specific model $\theta_{i'}$ is initialized by $\theta_0^*$, and is adapted to the new task $\mathcal{T}_{i'}$ by taking $K$ gradient steps on a small set of instances (different from test instances) sampled from $\mathcal{T}_{i'}$. Then the solution is constructed using the adapted policy;
3) \emph{active search:} the model is adapted to each test instance by learning instance-dependent parameters. ~\citet{bello2017neural} first proposes the general active search that iteratively adjusts the model parameters with the objective of increasing the likelihood of constructing high-quality solutions for each instance. However, it is extremely computationally expensive. ~\citet{hottung2022efficient} proposes the efficient active search (EAS) by introducing extra instance-specific parameters (e.g., a MLP layer) for each test instance while fixing the original model parameters. 
% We use zero-shot and few-shot to evaluate the generalization on a group of test instances, and use EAS for generalization on each single instance (see Appendix \ref{app:benchmark}). 
We mainly consider the zero-shot and few-shot settings in our experiments (in Section \ref{exp}), and only use EAS when evaluating on benchmark instances (in Appendix \ref{app:benchmark}).

\begin{figure}[!t]
    % \vskip 0.1in
    \centering
    \includegraphics[width=0.48\columnwidth]{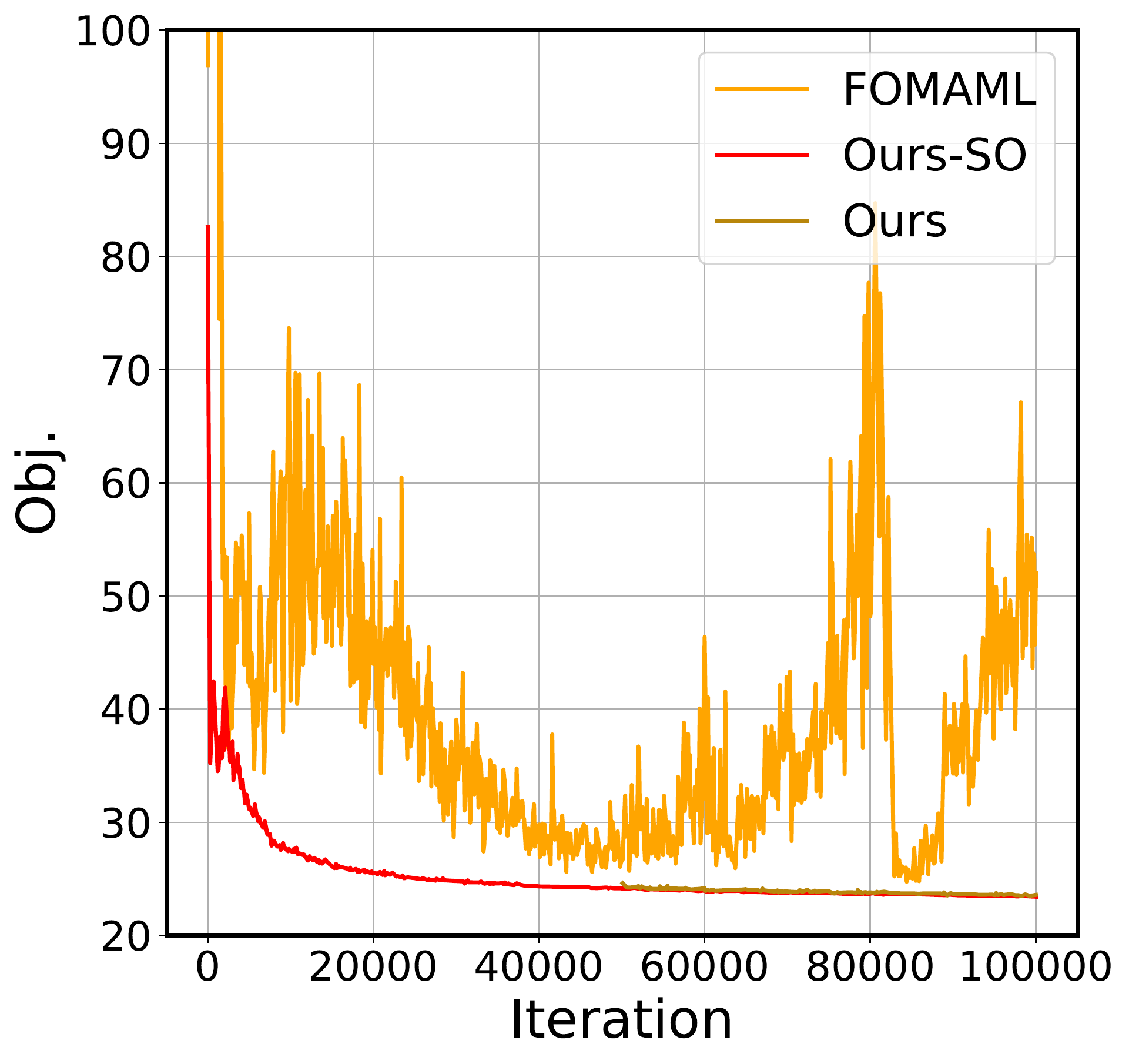}
    \includegraphics[width=0.495\columnwidth]{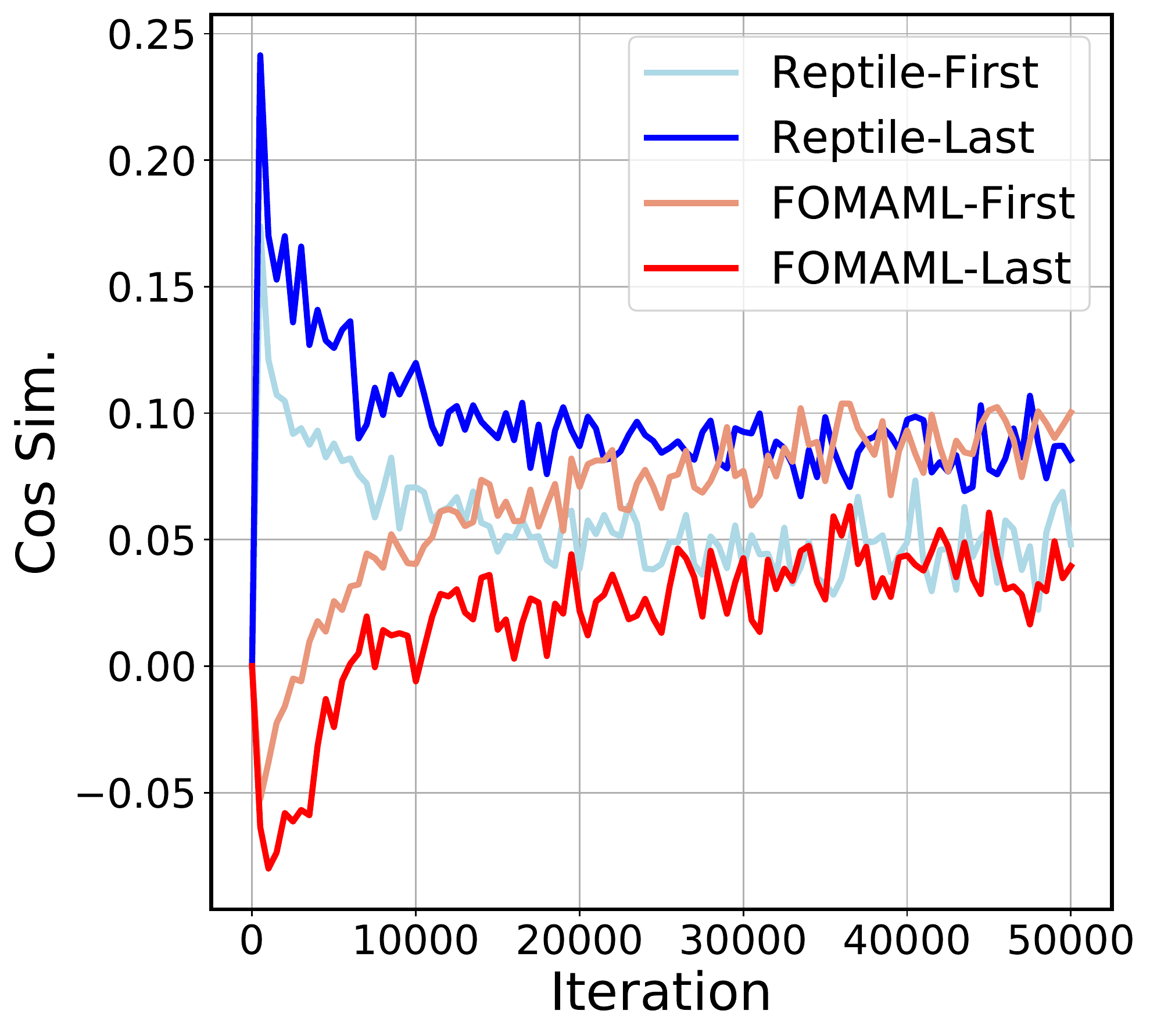}
    \vskip -0.05in
    \caption{\emph{Left panel:} the meta-training curves;
    \emph{Right panel:} the cosine similarity between the second-order derivative and its first-order approximation at the first and last layer of the model.
    }
    \label{story}
    \vskip -0.1in
\end{figure}

\subsection{First-Order Approximation}
The meta-gradient in Eq. (\ref{eq:meta_grad}) involves a gradient through a gradient (i.e., second-order derivative), and therefore it is computational expensive to obtain due to the calculation of Hessian-vector products.
% The meta-gradient in Eq. (\ref{eq:meta_grad}) is computationally expensive to obtain the second-order derivative.
%a gradient through \colorr{another} (i.e., second-order derivative), 
% and is computationally expensive due to the computation of Hessian matrices.
%Hessian-vector products. 
To tackle this issue, ~\citet{finn2017model} proposes a first-order approximation method (i.e., FOMAML) which simply drops the second-order term.
The empirical evidence of its effectiveness has been verified in the few-shot supervised learning, 
% They empirically observe the similar performance to MAML on the few-shot supervised learning setting, 
while lacking in the more complex reinforcement learning. Specifically, the first-order approximation of meta-model update can be expressed as:
\begin{equation}
    \label{eq:fomaml}
    \theta_0 \gets \theta_0- \beta \sum_{i=1}^{B} w_i\nabla_{\theta_i}\mathcal{L}_i(\theta_i^{(K)}).
\end{equation}
However, as shown in the left panel of Figure \ref{story}, we empirically observe that meta-training POMO from scratch with Eq. (\ref{eq:fomaml}) may induce fluctuating validation performance. The unstable meta-training may be attributed to the deviation of the first-order approximation (Eq. (\ref{eq:fomaml})) from the (ground-truth) gradient direction of the second-order term (line 16 in Algorithm \ref{alg:pomo-maml}). 
%The unstable meta-training may be attributed to the deviation from the (ground-truth) gradient direction of the second-order term, 
% at the \emph{early stage} of meta-training, 
% which is the only difference between the second-order method (line 16 in Algorithm \ref{alg:pomo-maml}) and the first-order approximation one (Eq. (\ref{eq:fomaml})).
Intuitively, sign$(\nabla_{\theta_i}\mathcal{L}_i(\theta_i^{(K)}))$ is the (steepest) descent direction for the task-specific model $\theta_i^{(K)}$, but not necessarily the descent direction for the meta-model $\theta_0$, especially at the \emph{early stage} of meta-training when the optimization tends to be unstable.
% Engineeringly, the initial oscillating reward signals, the choice of optimizer and normalization~\cite{antoniou2019how} may unstable the optimization as well.
To justify our hypothesis, we show the cosine similarities of the gradient directions for meta-model updates between our method and others in the right panel of Figure \ref{story}. The detailed experimental setups are presented in Appendix \ref{app:setup}. We observe that sign$(\nabla_{\theta_i}\mathcal{L}_i(\theta_i^{(K)}))$ cannot approximate sign$(\nabla_{\theta_0}\mathcal{L}_i(\theta_i^{(K)}))$ well (i.e., with a negative cosine similarity) at the early stage of meta-training, while gradually having the positive correlation as the training progresses.
Therefore, in order to reduce the computational cost and stabilize the meta-training, we develop a simple yet efficient method by early stopping the usage of second-order derivatives. Specifically, we start the meta-model updates with second-order derivatives, and switch to the first-order ones (i.e., by replacing line 16 with Eq. (\ref{eq:fomaml})) when the optimization tends to be stable.
% Specifically, it starts with the second-order meta-training method, and switches to the first-order approximation one (i.e., by replacing line 16 with Eq. (\ref{eq:fomaml})) when the optimization tends to be stable.
% However, theoretical and empirical evidences are still missing in a more complex reinforcement learning setting. In this subsection, we provide some empirical analysis about the effect of first-order approximation.
Recently, ~\citet{manchanda2022generalization} leverages another first-order method called Reptile~\cite{nichol2018first}, 
% , which only relies on the first-order derivative as well, 
with the form of the meta-model update as follows:
\begin{equation}
    \label{eq:reptile}
    \theta_0 \gets \theta_0 + \beta \sum_{i=1}^B w_i(\theta_i^{(K)} - \theta_0).
\end{equation}
% For simplicity, we set $w_i=\frac{1}{B}$ for all tasks in Section \ref{} and Appendix \ref{}. 
% In this paper, we empirically show the effectiveness of FOMAML on our setting. Surprisingly, we find that only one step of inner-loop optimization could achieve superior performance over Reptile, which takes much larger number of steps. 
However, it needs multiple inner-loop updates to effectively incorporate information from higher-order derivatives of the loss function so as to achieve satisfactory performance (as shown in  Appendix \ref{app:analysis} and \ref{app:meta-pomo}), making it less sample efficient. Moreover, similar to~\citet{nichol2018first} which observes negative results after applying Reptile to the reinforcement learning setting, we also empirically find its weak performance (see Meta-POMO in Section \ref{exp}, Appendix \ref{app:meta-pomo} and \ref{app:benchmark}).
% its performance in the reinforcement learning setting is not strong (as shown in Section \ref{exp}), or even be negetive~\cite{nichol2018first}. 
In contrast, our method could achieve decent performance only running a single inner-loop update.

\begin{table*}[!t]
  % \vskip -0.1in
  \caption{Evaluation on cross-size or distribution generalization.
  (* marks results derived by open-sourced pretrained models.)}
  \label{exp_1}
  % \vskip 0.1in
  \begin{center}
  \begin{small}
%   \begin{sc}
  \renewcommand\arraystretch{1.26}
  \resizebox{\textwidth}{!}{ 
  \begin{tabular}{ll|cccccc|cccccc}
    \toprule
    \midrule
    \multicolumn{2}{c|}{\multirow{3}{*}{Method}} &
    \multicolumn{6}{c|}{\textbf{Cross-Distribution Generalization} (1K ins.)} & \multicolumn{6}{c}{\textbf{Cross-Size Generalization} (1K ins.)} \\
     & & \multicolumn{2}{c}{$(200, GM_2^5)$} & \multicolumn{2}{c}{$(200, R)$} & \multicolumn{2}{c|}{$(200, E)$} & \multicolumn{2}{c}{$(300, U)$} & \multicolumn{2}{c}{$(300, GM_3^{10})$} & \multicolumn{2}{c}{$(300, GM_7^{50})$} \\
     & & Obj. (Gap) & Time & Obj. (Gap) & Time & Obj. (Gap) & Time & Obj. (Gap) & Time & Obj. (Gap) & Time & Obj. (Gap) & Time \\
    \midrule
    \multirow{13}*{\rotatebox{90}{TSP}} & Concorde & 8.78 (0.00\%) & 0.6m & 8.20 (0.00\%) & 0.5m & 8.09 (0.00\%) & 0.5m & 12.95 (0.00\%) & 1.4m & 9.47 (0.00\%) & 1.2m & 5.63 (0.00\%) & 1.0m \\
    & LKH3 & 8.78 (0.00\%) & 3.3m & 8.20 (0.00\%) & 3.3m & 8.09 (0.00\%) & 3.5m & 12.95 (0.00\%) & 5.9m & 9.47 (0.00\%) & 12.5m & 5.63 (0.01\%) & 18.5m \\
    \cmidrule{2-14}
     & POMO* & 9.36 (6.67\%) & 0.5m & 8.41 (2.66\%) & 0.5m & 8.28 (2.35\%) & 0.5m & 13.82 (6.70\%) & 1.5m & 10.73 (13.38\%) & 1.5m & 6.54 (16.04\%) & 1.5m \\
     & AMDKD-POMO* & 9.05 (2.97\%) & 0.5m & 8.41 (2.57\%) & 0.5m & 8.30 (2.61\%) & 0.5m & 13.97 (7.83\%) & 1.5m & 10.25 (8.22\%) & 1.5m & 6.25 (11.00\%) & 1.5m \\
    \cmidrule{2-14}
     & POMO & \textbf{9.01 (2.56\%)} & 0.5m & 8.37 (2.13\%) & 0.5m & 8.24 (1.85\%) & 0.5m & 13.54 (4.51\%) & 1.5m & 9.88 (4.27\%) & 1.5m & 5.83 (3.46\%) & 1.5m \\
     % & \sout{DIMES} & \sout{9.94 (13.20\%)} & 3.6m & \sout{9.29 (13.34\%)} & 3.2m & \sout{9.36 (15.64\%)} & 3.2m & \sout{14.85 (14.63\%)} & 5.1m & \sout{10.98 (16.02\%)} & 4.9m & \sout{6.32 (12.24\%)} & 4.4m \\
     & AMDKD-POMO \iffalse308 epoch\fi & 9.10 (3.56\%) & 0.5m & 8.47 (3.32\%) & 0.5m & 8.38 (3.55\%) & 0.5m & 13.74 (6.08\%) & 1.5m & 9.97 (5.30\%) & 1.5m & 6.00 (6.44\%) & 1.5m \\
    %  & Meta-POMO & 9.02 (2.64\%) & 0.5m & 8.37 (2.12\%) & 0.5m & 8.24 (1.87\%) & 0.5m & 13.47 (3.95\%) & 1.5m & 9.87 (4.19\%) & 1.5m & 5.80 (2.85\%) & 1.5m \\
     & Meta-POMO & 9.03 (2.78\%) & 0.5m & 8.39 (2.31\%) & 0.5m & 8.25 (2.00\%) & 0.5m & 13.50 (4.23\%) & 1.5m & 9.89 (4.38\%) & 1.5m & 5.80 (2.94\%) & 1.5m \\
     & Ours-SO & 9.01 (2.59\%) & 0.5m & \textbf{8.36 (1.99\%)} & 0.5m & \textbf{8.23 (1.72\%)} & 0.5m & \textbf{13.37 (3.22\%)} & 1.5m & \textbf{9.82 (3.72\%)} & 1.5m & \textbf{5.78 (2.61\%)} & 1.5m \\
     & Ours & 9.02 (2.71\%) & 0.5m & 8.37 (2.14\%) & 0.5m & 8.24 (1.86\%) & 0.5m & 13.40 (3.42\%) & 1.5m & 9.84 (3.89\%) & 1.5m & 5.79 (2.75\%) & 1.5m \\
    %  & Ours & \colorb{9.02 (2.66\%)} & 0.5m & \colorb{8.37 (2.08\%)} & 0.5m & \colorb{8.24 (1.78\%)} & 0.5m & \colorb{13.40 (3.41\%)} & 1.5m & \colorb{9.84 (3.85\%)} & 1.5m & \colorb{5.79 (2.71\%)} & 1.5m \\
    \cmidrule{2-14}
     & Meta-POMO+FS ($K=1$) & 9.02 (2.74\%) & 2.0m & 8.38 (2.24\%) & 2.0m & 8.25 (1.92\%) & 2.0m & 13.46 (3.87\%) & 6.8m & 9.86 (4.11\%) & 6.8m & 5.78 (2.64\%) & 6.8m \\
     & Meta-POMO+FS ($K=10$) & 9.02 (2.67\%) & 15.7m & 8.38 (2.17\%) & 15.7m & 8.24 (1.83\%) & 15.7m & 13.42 (3.58\%) & 0.9h & 9.84 (3.91\%) & 0.9h & \textbf{5.77 (2.46\%)} & 0.9h \\
     & Ours-SO+FS ($K=1$) & \textbf{9.01 (2.53\%)} & 2.0m & \textbf{8.36 (1.95\%)} & 2.0m & \textbf{8.22 (1.63\%)} & 2.0m & \textbf{13.35 (3.05\%)} & 6.8m & \textbf{9.81 (3.57\%)} & 6.8m & \textbf{5.77 (2.47\%)} & 6.8m \\
     & Ours+FS ($K=1$) & 9.01 (2.60\%) & 2.0m & 8.37 (2.05\%) & 2.0m & 8.23 (1.74\%) & 2.0m & 13.37 (3.19\%) & 6.8m & 9.82 (3.69\%) & 6.8m & 5.78 (2.52\%) & 6.8m \\
    %  & Ours+FS ($K=10$) & 9.01 (2.54\%) & 15.7m & 8.36 (2.00\%) & 15.7m & 8.23 (1.65\%) & 15.7m & 13.35 (3.08\%) & 0.9h & 0 & 0.9h & 0 & 0.9h \\
    \midrule
    
    \multirow{13}*{\rotatebox{90}{CVRP}} & HGS & 18.89 (0.00\%) & \iffalse 44.0m\fi 0.7h & 19.36 (0.00\%) & \iffalse33.5m\fi 0.6h & 19.45 (0.00\%) & \iffalse29.9m\fi 0.5h & 25.61 (0.00\%) & 1.0h & 22.20 (0.00\%) & 1.6h & 22.11 (0.00\%) & \iffalse56.7m\fi 0.9h \\
    & LKH3 & 19.09 (1.06\%) & \iffalse33.5m\fi 0.6h & 19.55 (0.99\%) & \iffalse33.5m\fi 0.6h & 19.66 (1.04\%) & \iffalse31.4m\fi 0.5h & 25.97 (1.39\%) & \iffalse36.2m\fi 0.6h & 22.46 (1.19\%) & \iffalse41.0m\fi 0.7h & 22.24 (0.59\%) & \iffalse36.1m\fi 0.6h \\
    \cmidrule{2-14}
     & POMO* & 19.93 (5.60\%) & 0.6m & 20.45 (5.74\%) & 0.6m & 20.54 (5.69\%) & 0.6m & 28.72 (12.32\%) & 1.8m & 24.81 (12.00\%) & 1.8m & 24.33 (10.22\%) & 1.8m \\
     & AMDKD-POMO* & 20.29 (7.59\%) & 0.6m & 20.89 (8.07\%) & 0.6m & 20.96 (7.92\%) & 0.6m & 30.49 (19.17\%) & 1.8m & 25.65 (15.92\%) & 1.8m & 24.41 (10.60\%) & 1.8m \\
    \cmidrule{2-14}
     & POMO & 19.47 (3.12\%) & 0.6m & 19.99 (3.32\%) & 0.6m & 20.12 (3.49\%) & 0.6m & 27.07 (5.74\%) & 1.8m & 23.25 (4.80\%) & 1.8m & 22.80 (3.16\%) & 1.8m \\
     & AMDKD-POMO & 19.58 (3.69\%) & 0.6m & 20.07 (3.72\%) & 0.6m & 20.20 (3.93\%) & 0.6m & 26.94 (5.20\%) & 1.8m & 23.28 (4.92\%) & 1.8m & 22.82 (3.27\%) & 1.8m \\
     & Meta-POMO & 19.48 (3.19\%) & 0.6m & 20.01 (3.40\%) & 0.6m & 20.15 (3.65\%) & 0.6m & 26.87 (4.94\%) & 1.8m & 23.09 (4.07\%) & 1.8m & 22.75 (2.93\%) & 1.8m \\
     & Ours-SO & \textbf{19.38 (2.66\%)} & 0.6m & \textbf{19.91 (2.87\%)} & 0.6m & \textbf{20.05 (3.13\%)} & 0.6m & 26.67 (4.15\%) & 1.8m & \textbf{22.93 (3.33\%)} & 1.8m & \textbf{22.60 (2.23\%)} & 1.8m \\
     & Ours & 19.39 (2.69\%) & 0.6m & 19.91 (2.88\%) & 0.6m & 20.07 (3.21\%) & 0.6m & \textbf{26.65 (4.10\%)} & 1.8m & 22.93 (3.35\%) & 1.8m & 22.61 (2.27\%) & 1.8m \\
    \cmidrule{2-14}
     & Meta-POMO+FS ($K=1$) & 19.43 (2.92\%) & 2.4m & 19.96 (3.13\%) & 2.4m & 20.10 (3.39\%) & 2.4m & 26.71 (4.32\%) & 8.2m & 22.99 (3.64\%) & 8.2m & 22.70 (2.72\%) & 8.2m \\
     & Meta-POMO+FS ($K=10$) & 19.41 (2.83\%) & 18.8m & 19.94 (3.03\%) & 18.8m & 20.08 (3.27\%) & 18.8m & 26.65 (4.07\%) & 1.1h & 22.95 (3.43\%) & 1.1h & 22.67 (2.55\%) & 1.1h \\
     & Ours-SO+FS ($K=1$) & \textbf{19.38 (2.65\%)} & 2.4m & \textbf{19.90 (2.81\%)} & 2.4m & \textbf{20.03 (3.00\%)} & 2.4m & 26.61 (3.93\%) & 8.2m & \textbf{22.90 (3.21\%)} & 8.2m & \textbf{22.58 (2.16\%)} & 8.2m \\
     & Ours+FS ($K=1$) & 19.38 (2.66\%) & 2.4m & 19.90 (2.83\%) & 2.4m & 20.04 (3.05\%) & 2.4m & \textbf{26.61 (3.92\%)} & 8.2m & 22.91 (3.23\%) & 8.2m & 22.59 (2.20\%) & 8.2m \\
    %  & Ours+FS ($K=10$) & 19.38 (2.63\%) & 18.8m & 19.89 (2.80\%) & 18.8m & 20.02 (2.99\%) & 18.8m & 26.57 (3.79\%) & 1.1h & 22.89 (3.17\%) & 1.1h & 22.58 (2.15\%) & 1.1h \\
    \midrule
    \bottomrule
  \end{tabular}}
%   \end{sc}
  \end{small}
  \end{center}
  \vskip -0.1in
\end{table*}

\section{Experiments}
\label{exp}
To demonstrate the effectiveness of the proposed framework, we apply it to POMO\footnote{\url{https://github.com/yd-kwon/POMO}}~\cite{kwon2020pomo}, which is a strong construction-based neural method.
We consider two representative VRP problems (i.e., TSP and CVRP). The details of POMO are introduced in Appendix \ref{app:pomo}.
Moreover, we also evaluate the generalizability of our method on L2D~\cite{li2021learning} as shown in Section \ref{exp:gene} and Appendix \ref{app:generalizability}.
% We will make our source code and test datasets public for future research on the challenging omni-generalization setting.

\textbf{Baselines.} 1) \emph{Traditional VRP solvers:} we employ Concorde and LKH3~\cite{helsgaun2017extension} for solving TSP, and the hybrid genetic search (HGS)~\cite{vidal2022hybrid} and LKH3 for CVRP.
2) \emph{Neural methods:} we compare our method with POMO-based methods, including the original POMO~\cite{kwon2020pomo}, AMDKD-POMO~\cite{bi2022learning} and Meta-POMO~\cite{manchanda2022generalization} for TSP and CVRP. 
AMDKD-POMO is a recent method that improves the cross-distribution generalization of POMO using knowledge distillation. Meta-POMO uses Reptile to improve the generalization performance across both size and distribution. 
For a fair comparision, we re-train all methods following our training setups.
Note that the setting of Meta-POMO is the most relevant to ours.
As shown in Appendix \ref{app:meta-pomo}, we tune the key hyperparameters (i.e., $\beta$ and $K$) of Meta-POMO since we empirically find its straightforward adaptation to POMO (i.e., \emph{decaying $\beta$} in Table \ref{sup_exp_1}) performs poorly. 
% It may be attributed to below reasons: a) they only consider around 10 train tasks and randomly select tasks to train, therefore failing to deal with our more complex setting; b) further designs may be needed in order to be successfully adapted to POMO, since POMO inherently improves generalization upon AM, on which they originally build. 
We also show the results of their open-sourced pretrained models (i.e., POMO* and AMDKD-POMO*) in Table \ref{exp_1}, with the aim of demonstrating the severe generalization issue of current neural methods rather than the direct comparison. Specifically, POMO* is trained on instances with a fixed size and distribution (i.e., $n=100$ with the uniform distribution), and AMDKD-POMO* is adaptively distilling from teacher models trained on fixed-sized instances following different distributions (i.e., $n=100$ with the uniform, cluster and mixed distributions). 
% Moreover, we compare our method with POMO, GANCO~\cite{xin2022generative}, AMDKD-POMO and Meta-POMO on benchmark datasets in Appendix \ref{app:benchmark}. 
More implementation details are provided in Appendix \ref{app:setup}. 
 
\textbf{Training Setups.} We follow most of the setups in~\citet{kwon2020pomo}. For our method, Adam optimizer~\cite{kingma2015adam} is used in both inner-loop and outer-loop optimization, with the weight decay of $1e-6$. The step sizes (learning rates) are $\alpha\!=\!\beta\!=\!1e-4$, and decayed by 10 in the last 10\% iterations to achieve a faster convergence. 
The batch size is $M\!=\!64$ ($M\!=\!32$ for instances with sizes larger than 150). 
The training task set consists of hundreds of (i.e., 341) tasks, with diverse sizes $\mathcal{N}=[50, 200]$ and distributions (i.e., uniform ($U$) and gaussian mixture ($GM$) distributions). More details about the generation of training and test data are  presented in Appendix \ref{app:data_gen}.
Similar to ~\citet{finn2017model}, we simply set $B\!=\!K\!=\!1$ and empirically observe strong performance. 
As suggested by~\citet{kwon2020pomo}, most of the training is already completed by 200 epochs (i.e., 20M instances) for POMO. We give more instances due to our complicated problem setting. Specifically, we re-train all methods for roughly the same number of instances (i.e., 32M) sampled from our training task set. For example, we re-train POMO for roughly 500K iterations (i.e., gradient updates). For our method, one iteration of meta-training consists of a pair of inner-loop and out-loop optimization, which needs two batches of instances. Therefore, for a fair comparison, we train our method for roughly 250K iterations. 
For the hierarchical task scheduler, we set $\eta\!=\!1$ and $E_s\!=\!225\text{K}$. It evaluates the hardness and updates the weight of each task every 100 iterations. 
Due to the training efficiency, we regard meta-training with the first-order approximation (i.e., \emph{Ours}) as the default method, which uses the second-order derivatives in the first 50K iterations, and switch to the first-order ones afterwards.
In specific, the meta-training with full second-order derivatives (i.e., \emph{Ours-SO}) needs roughly 5 days and 53GB GPU memory for TSP (6 days and 71GB GPU memory for CVRP), while \emph{Ours} needs 2.5 days and 17GB GPU memory for TSP (3 days and 25GB GPU memory for CVRP).
% Tesla V100S GPU 
% Due to the same amount of gradient updates involved during training, our training time is roughly the same as the one needed by POMO.

\textbf{Inference Setups.} For all neural methods, we use the greedy rollout with x8 instance augmentations following~\citet{kwon2020pomo}. 
% For DIMES, we sample 4096 solutions in parallel following ~\cite{qiu2022dimes}. 
% Note that similar to ~\cite{kool2018attention,kwon2020pomo}, we only consider standard searching strategies such as greedy rollout and sampling, which could directly reflect the goodness of learned policy (or probability distribution). 
We report the average results over the test dataset containing 1K instances. The reported time is the total time to solve the entire test dataset. The reported gaps are computed with respect to the traditional VRP solvers (i.e., Concorde for TSP, and HGS for CVRP).
Specifically, we evaluate the effectiveness of our method on the \emph{zero-shot} and \emph{few-shot} (FS) settings. For the zero-shot setting, the trained model is directly used to construct the solutions. We further evaluate meta-learning based methods (i.e., Meta-POMO, Ours-SO and Ours) on the few-shot setting, where we fine-tune the meta-model for $K$ iterations only using extra 1K instances sampled from the test task (0.003\% of instances used for meta-training). The instances are augmented following~\citet{kwon2020pomo}. Note that the instances for fine-tuning are different from the test ones. The Adam optimizer is used with the learning rate of $\alpha\!=\!1e-5$ and the weight decay of $1e-6$. Moreover, we further combine our method with EAS~\cite{hottung2022efficient} when evaluating on benchmark instances (see Appendix \ref{app:benchmark}).
% In addition to the \emph{zero-shot} setting, we further evaluate meta-learning based methods (i.e., Meta-POMO and Ours) on two other settings: 1) \emph{few-shot} (i.e., +FS), where we fine-tune the meta-model for $5$ iterations only using extra 1000 instances sampled from the test task (0.003\% of the number of instances used during meta-training). Note that the instances for fine-tuning is different from test instances; 2) \emph{active search} (i.e., +EAS), where we adapt the meta-model to each test instance independently by learning instance-dependent parameters. It is computational expensive to apply efficient active search ~\cite{hottung2022efficient} to large instances, therefore we only use the EAS-lay (T=100) for the demonstration purpose.

\begin{table*}[!t]
%   \vskip -0.05in
  \caption{Evaluation on cross-size and distribution generalization.}
  \label{exp_2}
  % \vskip 0.1in
  \begin{center}
  \begin{small}
%   \begin{sc}
  \renewcommand\arraystretch{1.3}
  \resizebox{\textwidth}{!}{ 
  \begin{tabular}{ll|cccc|cccc|cccc}
    \toprule
    \midrule
    \multicolumn{2}{c|}{\multirow{3}{*}{Method}} &
    \multicolumn{12}{c}{\textbf{Cross-Size and Distribution Generalization} (1K ins.)} \\
     & & \multicolumn{2}{c}{$(300, R)$} & \multicolumn{2}{c|}{$(300, E)$} & \multicolumn{2}{c}{$(500, R)$} & \multicolumn{2}{c|}{$(500, E)$} & \multicolumn{2}{c}{$(1000, R)$} & \multicolumn{2}{c}{$(1000, E)$} \\
     & & Obj. (Gap) & Time & Obj. (Gap) & Time & Obj. (Gap) & Time & Obj. (Gap) & Time & Obj. (Gap) & Time & Obj. (Gap) & Time \\
    \midrule
    \multirow{11}*{\rotatebox{90}{TSP}} & Concorde & 9.79 (0.00\%) & 1.2m & 9.48 (0.00\%) & 1.5m & 12.39 (0.00\%) & 5.0m & 11.73 (0.00\%) & 5.8m & 17.09 (0.00\%) & \iffalse44.1m\fi 0.7h & 15.66 (0.00\%) & \iffalse53.8m\fi 0.9h \\
    & LKH3 & 9.79 (0.00\%) & 6.0m & 9.48 (0.00\%) & 6.8m & 12.39 (0.00\%) & 11.8m & 11.73 (0.00\%) & 13.8m & 17.09 (0.00\%) & \iffalse25.7m\fi 0.4h & 15.66 (0.00\%) & \iffalse28.0m\fi 0.5h \\
    \cmidrule{2-14}
     & POMO & 10.23 (4.43\%) & 1.5m & 9.88 (4.20\%) & 1.5m & 13.63 (10.00\%) & 6.0m & 12.89 (9.88\%) & 6.0m & 20.74 (21.38\%) & \iffalse46.2m\fi 0.8h & 18.94 (20.97\%) & \iffalse46.2m\fi 0.8h \\
     & AMDKD-POMO & 10.35 (5.69\%) & 1.5m & 10.06 (6.15\%) & 1.5m & 13.74 (10.85\%) & 6.0m & 13.08 (11.52\%) & 6.0m & 20.73 (21.25\%) & 0.8h & 19.08 (21.85\%) & 0.8h \\
     & Meta-POMO & 10.22 (4.37\%) & 1.5m & 9.87 (4.14\%) & 1.5m & 13.56 (9.41\%) & 6.0m & 12.84 (9.44\%) & 6.0m & 20.51 (19.97\%) & \iffalse46.2m\fi 0.8h & 18.77 (19.88\%) & \iffalse46.2m\fi 0.8h \\
     & Ours-SO & \textbf{10.14 (3.54\%)} & 1.5m & \textbf{9.78 (3.13\%)} & 1.5m & \textbf{13.39 (8.07\%)} & 6.0m & \textbf{12.64 (7.73\%)} & 6.0m & \textbf{20.37 (19.20\%)} & \iffalse46.2m\fi 0.8h & \textbf{18.59 (18.74\%)} & \iffalse46.2m\fi 0.8h \\
     & Ours & 10.16 (3.74\%) & 1.5m & 9.80 (3.35\%) & 1.5m & 13.42 (8.30\%) & 6.0m & 12.66 (7.90\%) & 6.0m & 20.40 (19.36\%) & \iffalse46.2m\fi 0.8h & 18.60 (18.80\%) & \iffalse46.2m\fi 0.8h \\
    %  & Ours & \colorb{10.15 (3.65\%)} & 1.5m & \colorb{9.79 (3.28\%)} & 1.5m & \colorb{13.41 (8.18\%)} & 6.0m & \colorb{12.65 (7.82\%)} & 6.0m & \colorb{20.37 (19.22\%)} & 46.2m & \colorb{18.59 (18.74\%)} & 46.2m \\
    \cmidrule{2-14}
     & Meta-POMO+FS ($K=1$) & 10.18 (3.96\%) & 6.8m & 9.83 (3.70\%) & 6.8m & 13.34 (7.60\%) & 0.5h & 12.63 (7.66\%) & 0.5h & \iffalse19.75 (15.51\%)\fi 19.58 (14.52\%) & 6.5h & \iffalse18.09 (15.53\%)\fi 17.92 (14.48\%) & 6.5h \\
     & Meta-POMO+FS ($K=10$) & 10.16 (3.69\%) & 0.9h & 9.80 (3.41\%) & 0.9h & 13.23 (6.75\%) \iffalse13.20 (6.47\%)\fi & 4.1h & 12.54 (6.84\%) \iffalse12.49 (6.41\%)\fi & 4.1h & -- & -- & -- & -- \\
     & Ours-SO+FS ($K=1$) & \textbf{10.12 (3.32\%)} & 6.8m & \textbf{9.76 (2.91\%)} & 6.8m & \textbf{13.19 (6.45\%)} & 0.5h & \textbf{12.45 (6.11\%)} & 0.5h & \textbf{19.53 (14.28\%)} & 6.5h & 17.79 (13.65\%) & 6.5h \\
     & Ours+FS ($K=1$) & 10.13 (3.41\%) & 6.8m & 9.77 (3.05\%) & 6.8m & 13.20 (6.52\%) & 0.5h & 12.51 (6.64\%) & 0.5h & 19.53 (14.30\%) & 6.5h & \textbf{17.75 (13.38\%)} & 6.5h \\
    \midrule
    
    \multirow{11}*{\rotatebox{90}{CVRP}} & HGS & 22.40 (0.00\%) & 1.3h & 23.02 (0.00\%) & 1.3h & 26.62 (0.00\%) & 4.5h & 26.89 (0.00\%) & 4.6h & 32.36 (0.00\%) & 30.9h & 32.01 (0.00\%) & 37.7h \\
    & LKH3 & 22.68 (1.28\%) & \iffalse39.4m\fi 0.7h & 23.32 (1.28\%) & \iffalse38.7m\fi 0.7h & 27.06 (1.69\%) & \iffalse53.9m\fi 0.9h & 27.32 (1.61\%) & \iffalse54.6m\fi 0.9h & 33.16 (2.51\%) & 1.6h & 32.78 (2.43\%) & 1.6h \\
    \cmidrule{2-14}
     & POMO & 23.56 (5.30\%) & 1.8m & 24.20 (5.30\%) & 1.8m & 29.06 (9.48\%) & 6.9m & 29.29 (9.29\%) & 6.9m & 39.33 (22.44\%) & 1.0h & 38.63 (21.73\%) & 1.0h \\
     & AMDKD-POMO & 23.54 (5.18\%) & 1.8m & 24.24 (5.39\%) & 1.8m & 29.06 (9.32\%) & 6.9m & 29.33 (9.29\%) & 6.9m & 39.72 (23.17\%) & 1.0h & 38.86 (21.90\%) & 1.0h \\
     & Meta-POMO & 23.39 (4.54\%) & 1.8m & 24.08 (4.71\%) & 1.8m & 28.53 (7.34\%) & 6.9m & 28.80 (7.32\%) & 6.9m & 37.46 (16.09\%) & 0.9h & 36.85 (15.52\%) & 0.9h \\
     & Ours-SO & 23.24 (3.83\%) & 1.8m & \textbf{23.93 (4.07\%)} & 1.8m & 28.34 (6.60\%) & 6.7m & 28.63 (6.69\%) & 6.7m & 37.30 (15.62\%) & \iffalse51.1m\fi 0.8h & 36.61 (14.83\%) & \iffalse51.3m\fi 0.8h \\
     & Ours & \textbf{23.23 (3.79\%)} & 1.8m & 23.94 (4.08\%) & 1.8m & \textbf{28.29 (6.41\%)} & 6.7m & \textbf{28.60 (6.56\%)} & 6.7m & \textbf{37.02 (14.73\%)} & \iffalse51.1m\fi 0.8h & \textbf{36.40 (14.15\%)} & \iffalse51.3m\fi 0.8h \\
    \cmidrule{2-14}
     & Meta-POMO+FS ($K=1$) & 23.29 (4.05\%) & 8.2m & 23.96 (4.20\%) & 8.2m & 28.13 (5.80\%) & 0.6h & 28.43 (5.90\%) & 0.6h & 36.14 (11.93\%) & 7.5h & 35.78 (12.07\%) & 7.5h \\
     & Meta-POMO+FS ($K=10$) & 23.23 (3.79\%) & 1.1h & 23.90 (3.92\%) & 1.1h & \textbf{27.95 (5.14\%)} & 4.9h & \textbf{28.24 (5.19\%)} & 4.7h & -- & -- & -- & -- \\
     & Ours-SO+FS ($K=1$) & 23.19 (3.61\%) & 8.2m & \textbf{23.87 (3.78\%)} & 8.2m & 28.03 (5.41\%) & 0.6h & 28.33 (5.52\%) & 0.6h & 35.69 (10.52\%) & 7.4h & 35.40 (10.92\%) & 7.4h \\
     & Ours+FS ($K=1$) & \textbf{23.19 (3.59\%)} & 8.2m & 23.87 (3.79\%) & 8.2m & \textbf{28.01 (5.34\%)} & 0.6h & \textbf{28.31 (5.44\%)} & 0.6h & \textbf{35.60 (10.26\%)} & 7.4h & \textbf{35.25 (10.45\%)} & 7.4h \\
    \midrule
    \bottomrule
  \end{tabular}}
%   \end{sc}
  \end{small}
  \end{center}
  \vskip -0.1in
\end{table*}

\subsection{Performance Evaluation}
\label{exp:eval}
Below, we demonstrate the effectiveness of our method on synthetic and real-world datasets. 
For the synthetic data, we evaluate the generalization performance across size, distribution and the both.
%further split it into two different settings as shown in Table \ref{exp_1} and \ref{exp_2}. 
Note that we conduct t-test (with threshold of $5\%$) to verify the statistical significance, if the average objectives of two neural methods are close. 
%We observe that the null hypothesis is rejected, demonstrating that the reported results are statistically significant.

\textbf{Cross-Size or Distribution Generalization.}
We first consider a simple setting where either the cross-size or distribution generalization is evaluated. For the cross-distribution setting, we test on instances of size $n\!=\!200\!\in\![50, 200]$, while following diverse distributions 
% (e.g., $GM_2^5$ with the cluster $c=2$ and scale $l=5$) 
that are unseen during training.
Note that we do not strictly choose the test tasks sampled from the presumed training task distribution $p(\mathcal{T})$ since it only covers a small part of the entire problem space. 
% A complete and efficient data generator for COPs is still an open problem~\cite{yehuda2020s}. 
Therefore, we also evaluate all methods on several complex distributions, e.g., rotation ($R$) and explosion ($E$) distributions~\cite{bossek2019evolving}. For the cross-size setting, we evaluate on instances of the size $n=300 \notin [50, 200]$ following distributions used in training. 
Besides the zero-shot setting, we further compare with another meta-learning based method (i.e., Meta-POMO) on the few-shot setting, where we fine-tune the learned model for $K$ steps only using limited data. The detailed results are shown in Table \ref{exp_1}, where we observe that our method can achieve superior performance on both settings. 
The inferior performance of AMDKD-POMO may be attributed to its design for the trivial problem setting and sample inefficiency. While it is specialized for the cross-distribution generalization, its original problem setting is much easier than ours, with only three distributions on the fixed size (i.e., 100) considered during training. To achieve satisfactory performance, a good pretrained model for each training task is needed, which requires a huge amount of training instances.
% Note that we also show the results of the open-sourced pretrained models in Table \ref{exp_1}, which aim to demonstrate the severe generalization issue of current neural methods rather than direct comparison. Specifically, POMO* is trained on instances with a fixed size and distribution (i.e., $n=100$ with the uniform distribution), and AMDKD-POMO* only deal with the cross-distribution generalization by adaptively distilling from student models trained on fixed-sized instances following different distributions (i.e., $n=100$ with the uniform, cluster and mixed distributions). 

\textbf{Cross-Size and Distribution Generalization.}
We further evaluate all methods on a much more complex setting, where the generalization across both size and distribution is considered. Specifically, we choose the test task with the unseen size $n\!\in\! [300, 500, 1000]$ and distribution $d\!\in\! [R, E]$ during training. The results are presented in Table \ref{exp_2}, where we observe our method has consistently better performance than baselines. Notably, our method achieves superior results on the large-scale CVRP1000 task with totally unseen distributions, showing a strong omni-generalization capability.

\textbf{Results on Benchmark Datasets.}
We further evaluate all methods on the well-konwn benchmark datasets TSPLIB~\cite{reinelt1991tsplib} and CVRPLIB (Set-X~\cite{uchoa2017new} and Set-XML100~\cite{queiroga202210}). Detailed results can be found in Appendix \ref{app:benchmark}, where we observe our method performs well in most cases.

\subsection{Analyses}
\label{exp:ana}
In this section, we conduct further analyses, including the ablation studies and few-shot experiments, to demonstrate the effectiveness and sensitivity of the proposed framework.
More ablation studies on hyperparameters, optimizers and normalization layers are presented in Appendix \ref{app:ablation}.

\textbf{Ablation Study on Components.}
In Section \ref{exp:eval}, 
we have shown the effect of the first-order approximation. Compared with the full second-order method, it could achieve similar or even better zero-shot and few-shot performance, and meanwhile greatly reduce the training complexity. Here, following the training setups presented in Section \ref{exp}, we further conduct the ablation study on CVRP to demonstrate the benefit of each component in our framework. The results are shown in Table \ref{exp_abl}, where we observe that 
% which reveals the contribution of the two components to the generalization performance.
the meta-training significantly improves POMO (zero-shot) performance on the large-scale instances, and the task scheduler can further boost the overall performance of the meta-training.

\begin{table}[!t]
  \vskip -0.05in
  \caption{Ablation study on Components.}
  \label{exp_abl}
  \vskip 0.1in
  \begin{center}
  % \begin{small}
  \resizebox{0.45\textwidth}{!}{ 
  \begin{tabular}{l|cccc}
    \toprule
    \midrule
      & $(200,GM_2^5)$ & $(300,U)$ & $(500,R)$ & $(1000,E)$ \\
    %  & Obj. & Gap & Obj. & Gap \\
    \midrule
    POMO & 3.12\% & 5.74\% & 9.48\% & 21.73\% \\
    + task scheduler & \textbf{2.67\%} & 4.44\% & 7.53\% & 19.03\% \\
    + meta-training & 3.08\% & 4.79\% & 7.02\% & 15.39\% \\
    Ours & 2.69\% & \textbf{4.10\%} & \textbf{6.41\%} & \textbf{14.15\%} \\
    % W/O task scheduler & 3.08\% & 4.79\% & 7.02\% & 15.39\% \\
    % W/O meta-training & \textbf{2.67\%} & 4.44\% & 7.53\% & 19.03\% \\
    \midrule
    \bottomrule
  \end{tabular}}
%   \end{sc}
  % \end{small}
  \end{center}
  \vskip -0.15in
\end{table}

\begin{figure}[!ht]
    % \vskip -0.1in
    \centering
    \includegraphics[width=0.45\columnwidth]{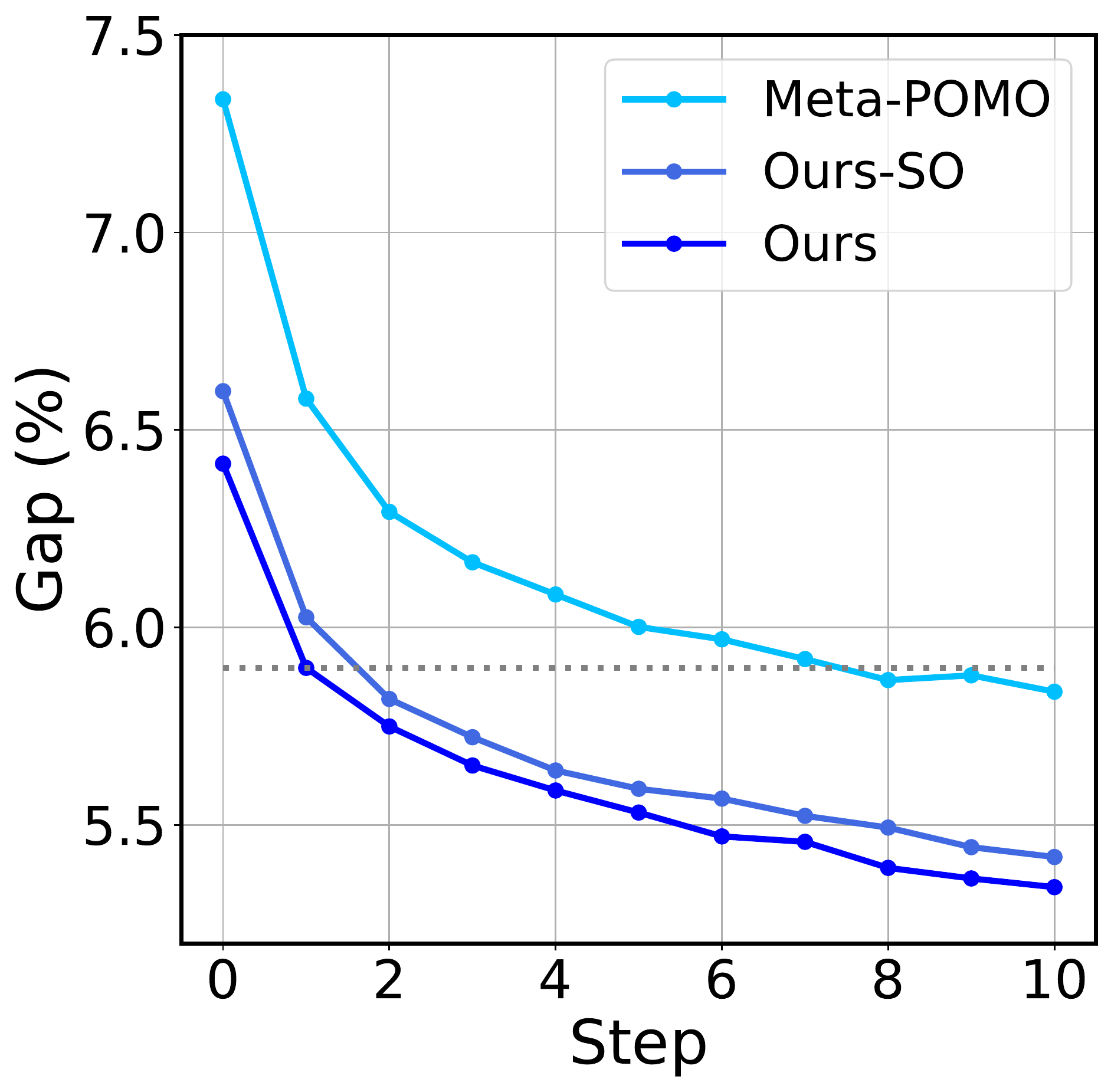}
    \includegraphics[width=0.45\columnwidth]{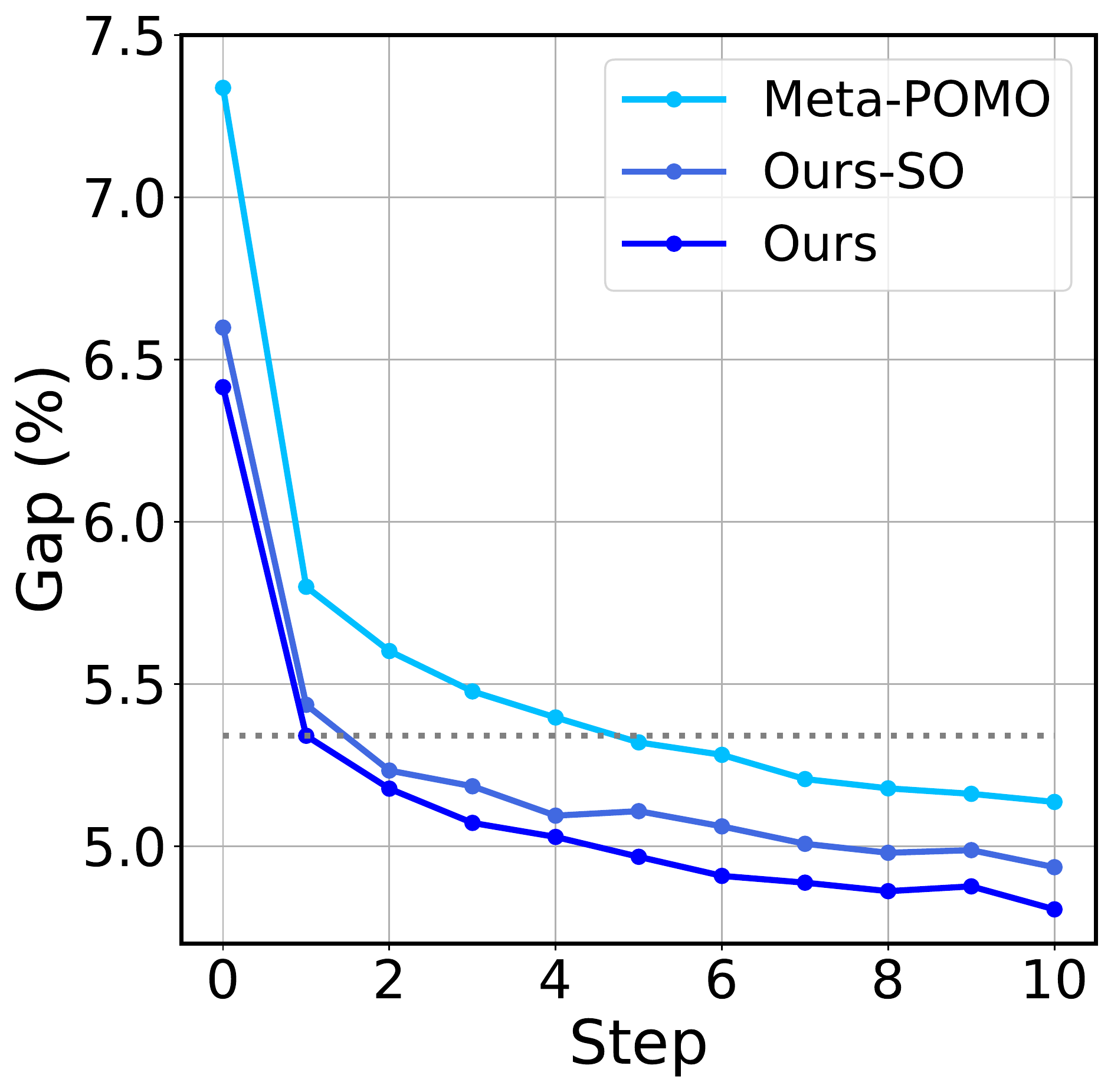}
    \vskip -0.05in
    \caption{Adaptation to CVRP (500, $R$) test task using (a) 100 instances; (b) 1000 instances.
    }
    \label{exp_3}
    \vskip -0.15in
\end{figure}

\textbf{Efficient Adaptation.}
As shown in Table \ref{exp_1} and \ref{exp_2}, given the same amount of training instances, our method achieves strong zero-shot performance, which enables efficient adaptation to a new task afterwards. 
To demonstrate it, we further conduct two experiments on the adaptation to CVRP (500, $R$) only using 100 and 1000 instances. 
% We conduct two more experiments on CVRP to demonstrate it. As shown in Figure \ref{exp_3}, 
As shown in Figure \ref{exp_3}, our methods can be efficiently adapted to the new task, while Meta-POMO needs to run multiple steps to achieve similar few-shot performance (e.g., the dotted line for $K=1$) to our method.
%, due to its inferior zero-shot performance.
% \colorb{Although Meta-POMO have a relative large slope at the last several steps, its few-shot performance is not strong due to the inferior zero-shot performance}. 
% However, due to the sample inefficiency of Meta-POMO during training, it cannot achieve the zero-shot performance of our method even taking more fine-tuning steps. 
Moreover, we observe that the number of instances is crucial to the few-shot performance in the reinforcement learning. Meta-POMO may need more instances in order to achieve strong few-shot performance.

% \newpage
\subsection{Generalizability}
\label{exp:gene}
To evaluate the generalizability of the proposed framework, we further apply it to L2D~\cite{li2021learning}, which is an improvement-based method outperforming LKH3 on large-scale CVRP instances. 
Specifically, it decomposes the large-scale problem instance into several subproblems, which are selected by a (supervised) learned policy, and uses an existing solver (e.g., LKH3 or HGS) to solve each subproblem. 
We train the model on the omni-generalization setting, where the training task set consists of various sizes and distributions. The results show that our method could improve the generalization of L2D, demonstrating the effectiveness and generalizability of our method.
The detailed training setups and empirical results are shown in Appendix \ref{app:generalizability}.

\section{Conclusion}
\label{conclusion}
% \colorb{learning an efficient adaptive task scheduler or sample scheduler within each task; learning several meta-models}
%To the best of our knowledge, this is the early work to study the challenging yet realistic generalization issue across both size and distribution in VRPs. 
% This paper aims to develop omni-generalizable neural methods across both problem size and distribution for solving VRPs. We propose a generic meta-learning framework, which is model-agnostic and compatible with any model trained with gradient updates, to achieve this goal.
%the unified generalization issue.
% which learns a good initialized model by meta-training tasks adaptively selected via a hierarchical task scheduler. The learned meta-model is able to be fast adapted to new tasks within a few steps of gradient descent during inference. 
% We further greatly improve the meta-training efficiency by applying first-order approximation methods on our RL setting.
% In this paper, we focus on the general meta-learning algorithm, which is model-agnostic and compatible with any model trained with gradient updates. 
% We will apply more advanced meta-learning algorithms in our framework (e.g., MAML with bootstrapping~\cite{flennerhag2022bootstrapped}) and improve other neural VRP methods (e.g., neural improvement heuristics). 
This paper studies the omni-generalization issue of neural methods across both problem size and distribution in VRPs. We propose a generic meta-learning framework to tackle this issue, which is model-agnostic and compatible with any model trained with gradient updates.
We further provide analyses of the first-order approximation methods on the reinforcement learning setting, and propose a simple yet efficient method to reduce the meta-training complexity.

The limitations of this work are the training efficiency and scalability. However, they heavily depend on the base model and meta-learning algorithm. If a pretrained model exists, it would be better to conduct meta-training on it. We refer to Appendix \ref{app:discussion} for further discussions. 
% Potential future work may include: 1) develop more advanced (e.g., MAML with bootstrapping~\cite{flennerhag2022bootstrapped}) or specialized meta-learning algorithms for VRPs; 2) reduc
We leave advanced algorithms and other neural VRP methods to the future work. 
We hope our work could provide new insights for learning a more generalizable neural VRP heuristic in the community.

% Acknowledgements should only appear in the accepted version.
\section*{Acknowledgements}
Wen Song was supported by the National Natural Science Foundation of China under Grant 62102228, and the Natural Science Foundation of Shandong Province under Grant ZR2021QF063.
We would like to thank the anonymous reviewers and (S)ACs of ICML 2023 for their constructive comments and dedicated service to the community. Jianan Zhou would like to personally express deep gratitude to his grandmother, Zhiling Kang, for her meticulous care and love during last 25 years. Eternal easy rest in sweet slumber.

% In the unusual situation where you want a paper to appear in the
% references without citing it in the main text, use \nocite
% \nocite{langley00}
\bibliographystyle{icml2023}
\bibliography{camera_ready}

%%%%%%%%%%%%%%%%%%%%%%%%%%%%%%%%%%%%%%%%%%%%%%%%%%%%%%%%%%%%%%%%%%%%%%%%%%%%%%%
%%%%%%%%%%%%%%%%%%%%%%%%%%%%%%%%%%%%%%%%%%%%%%%%%%%%%%%%%%%%%%%%%%%%%%%%%%%%%%%
% APPENDIX
%%%%%%%%%%%%%%%%%%%%%%%%%%%%%%%%%%%%%%%%%%%%%%%%%%%%%%%%%%%%%%%%%%%%%%%%%%%%%%%
%%%%%%%%%%%%%%%%%%%%%%%%%%%%%%%%%%%%%%%%%%%%%%%%%%%%%%%%%%%%%%%%%%%%%%%%%%%%%%%
\newpage
\appendix
\onecolumn

\section{Analysis of First-Order Approximation}
\label{app:analysis}
\begin{figure}[!t]
    \vskip 0.1in
    \centering
    \subfigure[First layer]{
      \includegraphics[width=0.23\columnwidth]{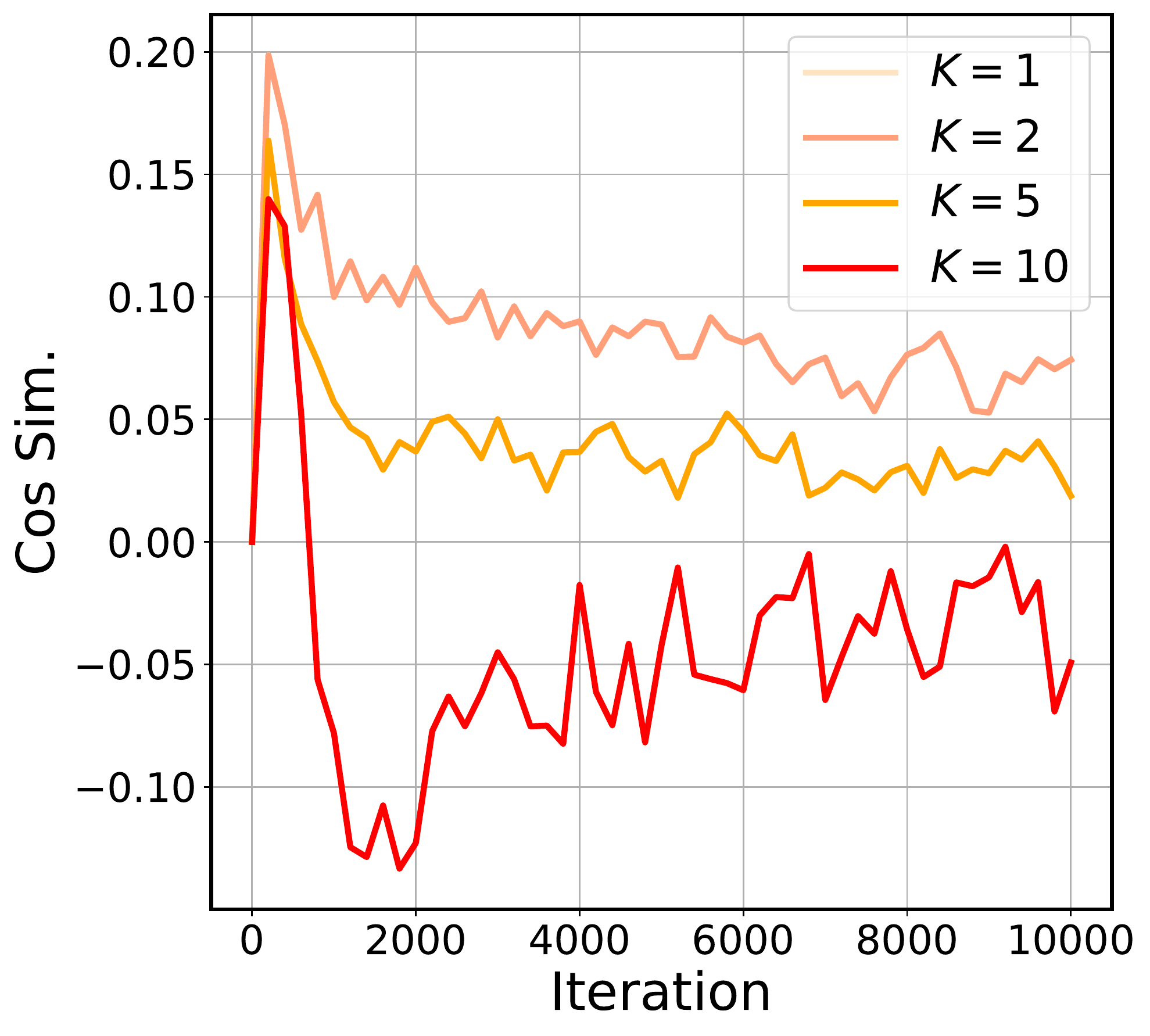}
    \label{story_a}}
    \subfigure[Last layer]{
      \includegraphics[width=0.23\columnwidth]{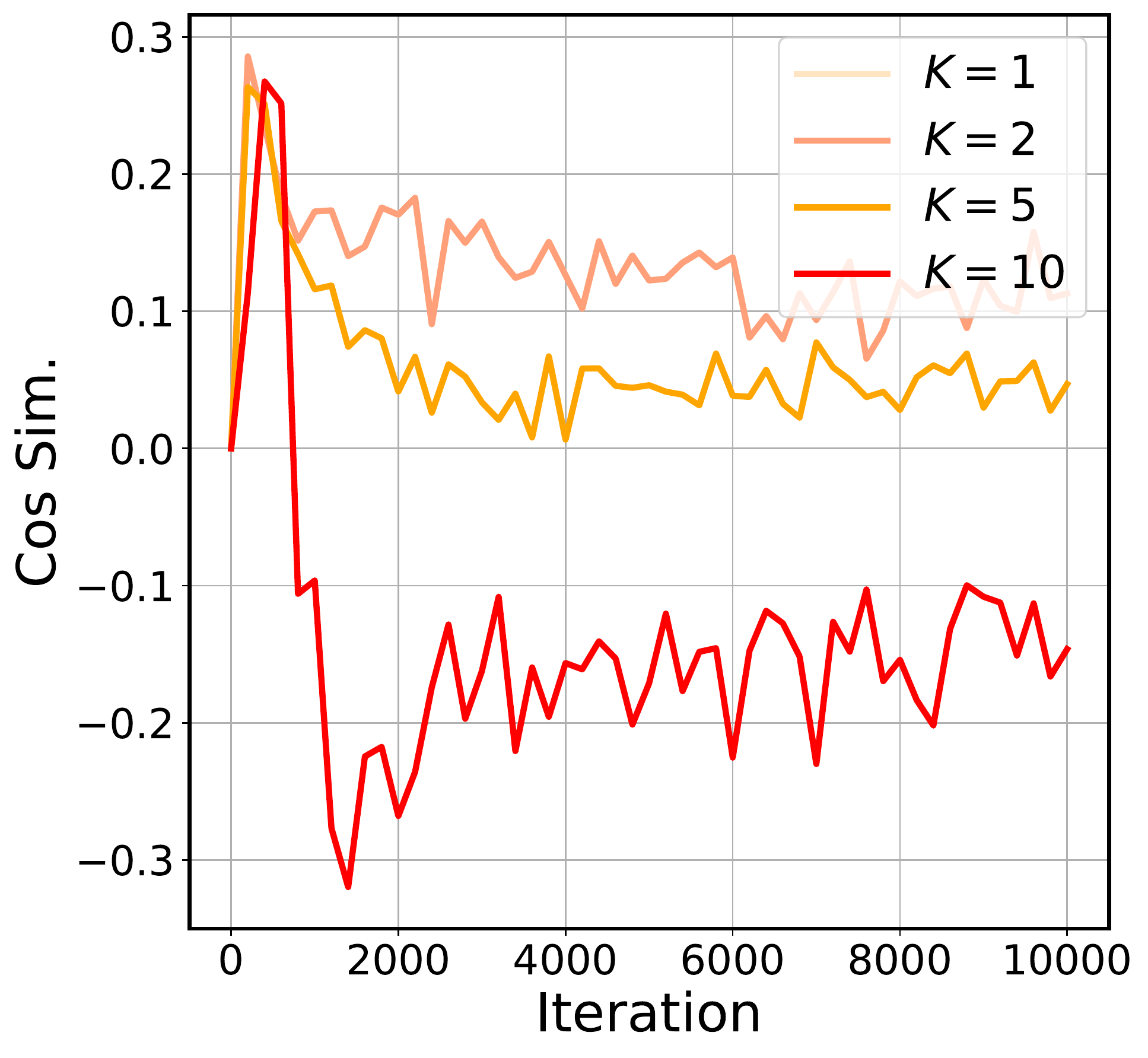}
    \label{story_b}}
    \subfigure[First layer]{
      \includegraphics[width=0.23\columnwidth]{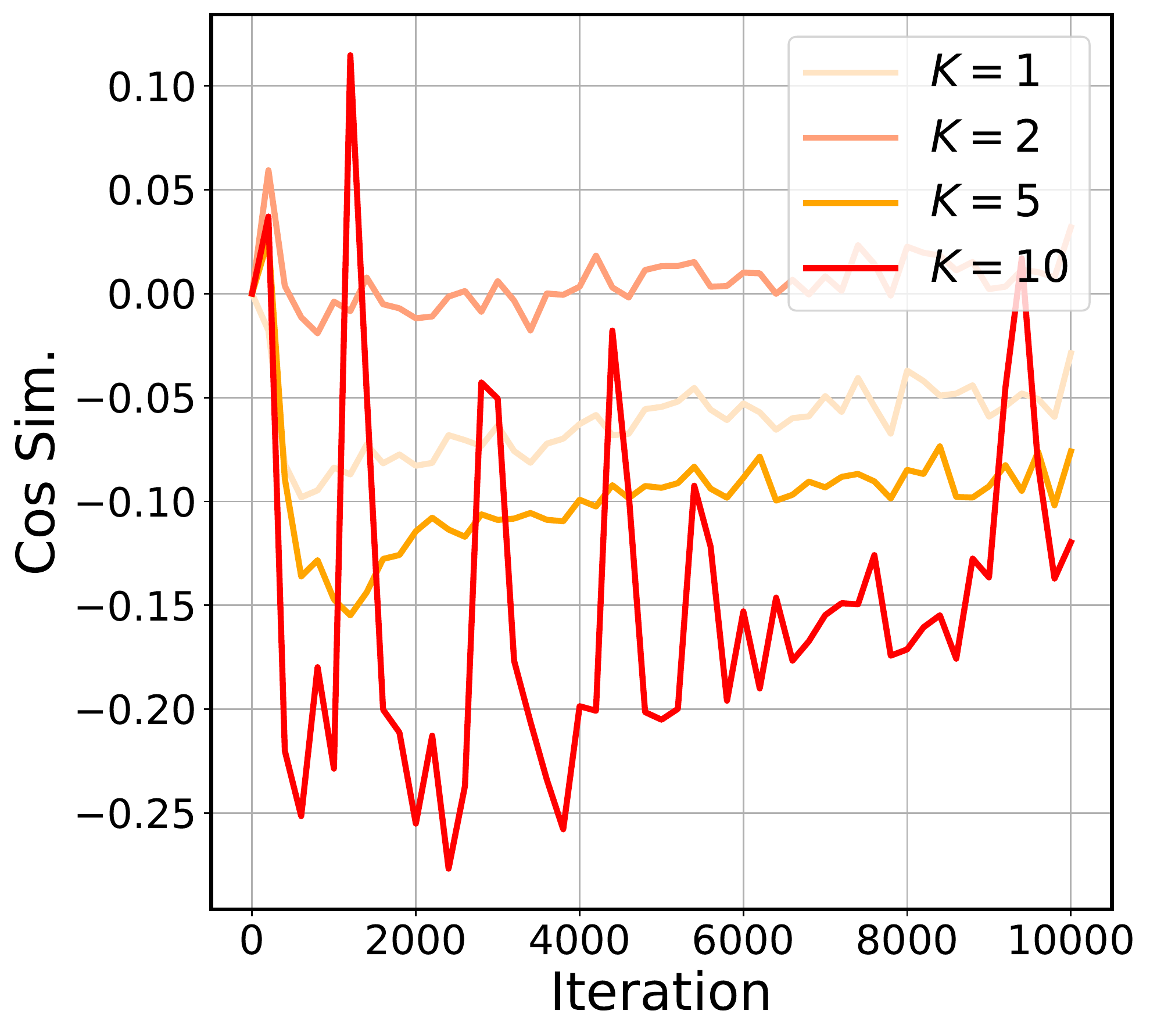}
    \label{story_c}}
    \subfigure[Last layer]{
      \includegraphics[width=0.23\columnwidth]{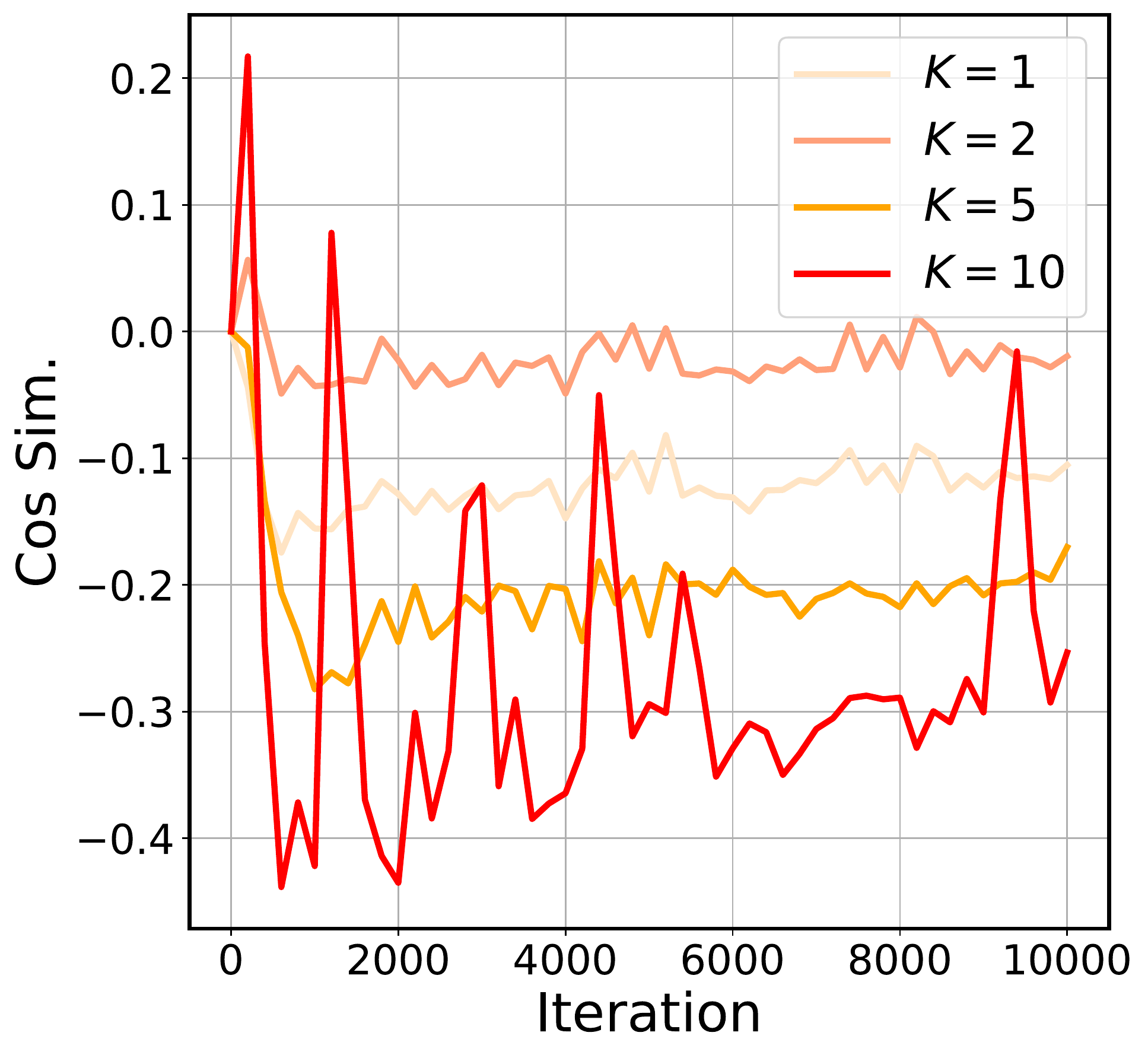}
    \label{story_d}}
    % \subfigure[Adaptation curve]{
    %   \includegraphics[width=0.185\columnwidth]{imgs/load_last.pdf}
    % \label{story_e}}
    % \vskip -0.05in
    \caption{(a)-(b): the cosine similarity of gradient directions between the second-order derivative and the Reptile's approximation when not loading gradient statistics. Note that the results of $K=1$ and $K=2$ are almost the same; (c)-(d): the cosine similarity of gradient directions between the second-order derivative and the Reptile's approximation when loading gradient statistics.
    }
    \label{app:story}
    % \vskip -0.1in
\end{figure}

Without loss of generality, we consider optimization with the stochastic gradient descent (SGD), and treat each task equally. Therefore, the meta-model update in Reptile~\cite{nichol2018first} (i.e., Eq. (\ref{eq:reptile})) could be rewritten as:
\begin{equation}
    \theta_0 \gets \theta_0 + \beta \frac{1}{B}\sum_{i=1}^B (\theta_i^{(K)} - \theta_0).
\end{equation}
For each task $\mathcal{T}_i$, $(\theta_0-\theta_{i}^{(K)})/\alpha$ could be viewed as the (Reptile) gradient term $g_R$ in the SGD formulation,
% ($\theta_0 \gets \theta_0-\beta g_R$), 
and $\alpha, \beta$ are the step sizes of inner-loop and outer-loop optimization, respectively. When $K=1$, Reptile is equivalent to the joint training on the expected loss of the training tasks:
\begin{equation}
    g_R^{1} = \mathbb{E}_{\mathcal{T}_i \sim p(\mathcal{T})} [\frac{ \theta_0-\theta_{i}^{(1)}}{\alpha}] = \mathbb{E}_{\mathcal{T}_i \sim p(\mathcal{T})} [\frac{\theta_0 - (\theta_0-\alpha \nabla_{\theta_0} \mathcal{L}_i(\theta_0))}{\alpha}] = \mathbb{E}_{\mathcal{T}_i \sim p(\mathcal{T})} [\nabla_{\theta_0} \mathcal{L}_i(\theta_0)].
\end{equation}
However, in this case, it is equivalent to the naive pretraining on a large training task set, which requires ad-hoc tricks to achieve desirable fine-tuning performance. When performing multiple gradient updates ($K>1$) in the inner-loop optimization, Reptile is able to incorporate information from higher-order derivatives of the loss function.
% In order to incorporate information from higher-order derivatives of the loss function, it needs to perform multiple gradient updates ($K>1$) in the inner-loop optimization. 
For the simplicity of notations, we omit the index for task $i$, and use the following definitions:
% \begin{align}
%     g^{(k)} &= \frac{\partial \mathcal{L}(\theta^{(k)})}{\partial \theta^{(k)}} \\
%     \tilde{g}^{(k)} &= \frac{\partial \mathcal{L}(\theta^{(k)})}{\partial \theta^{(0)}} \\
%     h^{(k)} &= \frac{\partial \mathcal{L}^2 (\theta^{(k)})} {\partial (\theta^{(0)})^2} 
% \end{align}
\begin{equation}
    g^{(k)} = \frac{\partial \mathcal{L}(\theta^{(k)})}{\partial \theta^{(k)}};\quad 
    \tilde{g}^{(k)} = \frac{\partial \mathcal{L}(\theta^{(k)})}{\partial \theta^{(0)}};\quad 
    \tilde{h}^{(k)} = \frac{\partial \mathcal{L}^2 (\theta^{(k)})} {\partial (\theta^{(0)})^2};\quad k \in [0, K],
\end{equation}
where $g^{(k)}, \tilde{g}^{(k)}$ are the gradients of the loss function with respect to (w.r.t.) the task-specific model $\theta^{(k)}$ and meta-model $\theta^{(0)}=\theta_0$, and $\tilde{h}^{(k)}$ is the hessian w.r.t. the meta-model. 
With the Taylor expansion, the gradient of the loss function w.r.t the task-specific model can be expressed as:
\begin{align}
  \label{eq:g}
  \begin{split}
    g^{(k)} = \frac{d\mathcal{L}}{d\theta}|_{\theta=\theta^{(k)}} &\approx \frac{d\mathcal{L}}{d\theta}|_{\theta=\theta^{(0)}} + \frac{d^2\mathcal{L}}{d\theta^2}|_{\theta=\theta^{(0)}} (\theta^{(k)} - \theta^{(0)}) \\
    &\approx \frac{\partial \mathcal{L}(\theta^{(k)})}{\partial \theta^{(0)}} + \frac{\partial \mathcal{L}^2(\theta^{(k)})}{\partial (\theta^{(0)})^2} (\theta^{(0)}-\alpha \sum_{j=0}^{k-1}\frac{\partial \mathcal{L}(\theta^{(j)})}{\partial \theta^{(j)}} - \theta^{(0)}) \\
    &\approx \tilde{g}^{(k)} - \alpha \tilde{h}^{(k)} \sum_{j=0}^{k-1}g^{(j)}.
  \end{split}
\end{align}
Indeed, FOMAML~\cite{finn2017model} simply drops the higher-order term and uses $g^{(K)}$ as the approximation to the second-order derivative, while Reptile approximates it in the following way:
% The gradient of Reptile for one task could be further written as:
\begin{equation}
  \label{eq:r}
    g_R^{K} = \frac{1}{\alpha} (\theta_0-\theta^{(K)}) = \frac{1}{\alpha} (\theta_0 - (\theta_0-\alpha\sum_{j=0}^{K-1}g^{(j)})) = \sum_{j=0}^{K-1}g^{(j)}.
\end{equation}
% Therefore, it needs multiple inner-loop updates to effectively incorporate information from higher-order derivatives of the loss function. 
For example, if we run $K=2$ steps in the inner-loop optimization, based on Eqs. (\ref{eq:g})-(\ref{eq:r}), the gradient of Reptile is $g_R^2 = g^{(0)} + g^{(1)} \approx \tilde{g}^{(0)} + \tilde{g}^{(1)} - \alpha \tilde{h}^{(1)}g^{(0)}$, and the gradient of FOMAML is $g_F^2 = g^{(2)} \approx \tilde{g}^{(2)} - \alpha \tilde{h}^{(2)}(g^{(0)}+g^{(1)})$. 
However, it is non-trivial to execute the above derivations on our more complicated reinforcement learning (RL) setting (with Adam optimizer). Therefore, empirically, we further conduct an experiment to check whether $g_R^{K}$ could serve as a good approximation on our setting. Specifically, similar to the setups presented in Appendix \ref{app:setup}, we meta-train POMO with Ours-SO for $K\in [1,2,5,10]$ steps in the inner-loop optimization. We collect the gradient direction of the second-order derivative sign$(\tilde{g}^{(K)})$ and that of the Reptile's approximation sign$(g_R^K)$, and compute their cosine similarity. Moreover, since we use the Adam optimizer, we also try to load the gradient statistics (e.g., momentum in the optimizer for outer-loop optimization) when conducting the inner-loop optimization. 
As indicated in Figure \ref{app:story}, Reptile fails to well approximate the second-order derivatives on our RL setting.
% which may be one of the reasons for its inferior performance. 
As the step $K$ increases, the cosine similarity decreases accordingly, which may be attributed to the accumulated effect throughout the $K$ steps. However, a larger step $K$ empirically results in a relatively better zero-shot generalization performance (given the same amount of training instances) as shown in Appendix \ref{app:meta-pomo}. 
% This counterintuitive evidence could be explained by the fact that the 
% 2) constructing an optimizer from scratch (rather than loading the previous gradient statistics) may benefits the few-shot generalization performance during the fine-tuning process.
% Moreover, different from FOMAML~\cite{finn2017model}, which is another first-order approximation method that simply drops the high-order term and uses $g^{(K)}$ as the approximation, Reptile does not need to split the data into training and validation sets.

\newpage
\section{Data Generation}
\label{app:data_gen}
% at least 0.1~inches of space before the caption and 0.1~inches after it
\begin{wrapfigure}{l}{0.5\columnwidth}
    \vskip 0.1in
    \begin{center}
    \subfigure[]{
    \includegraphics[width=0.12\columnwidth]{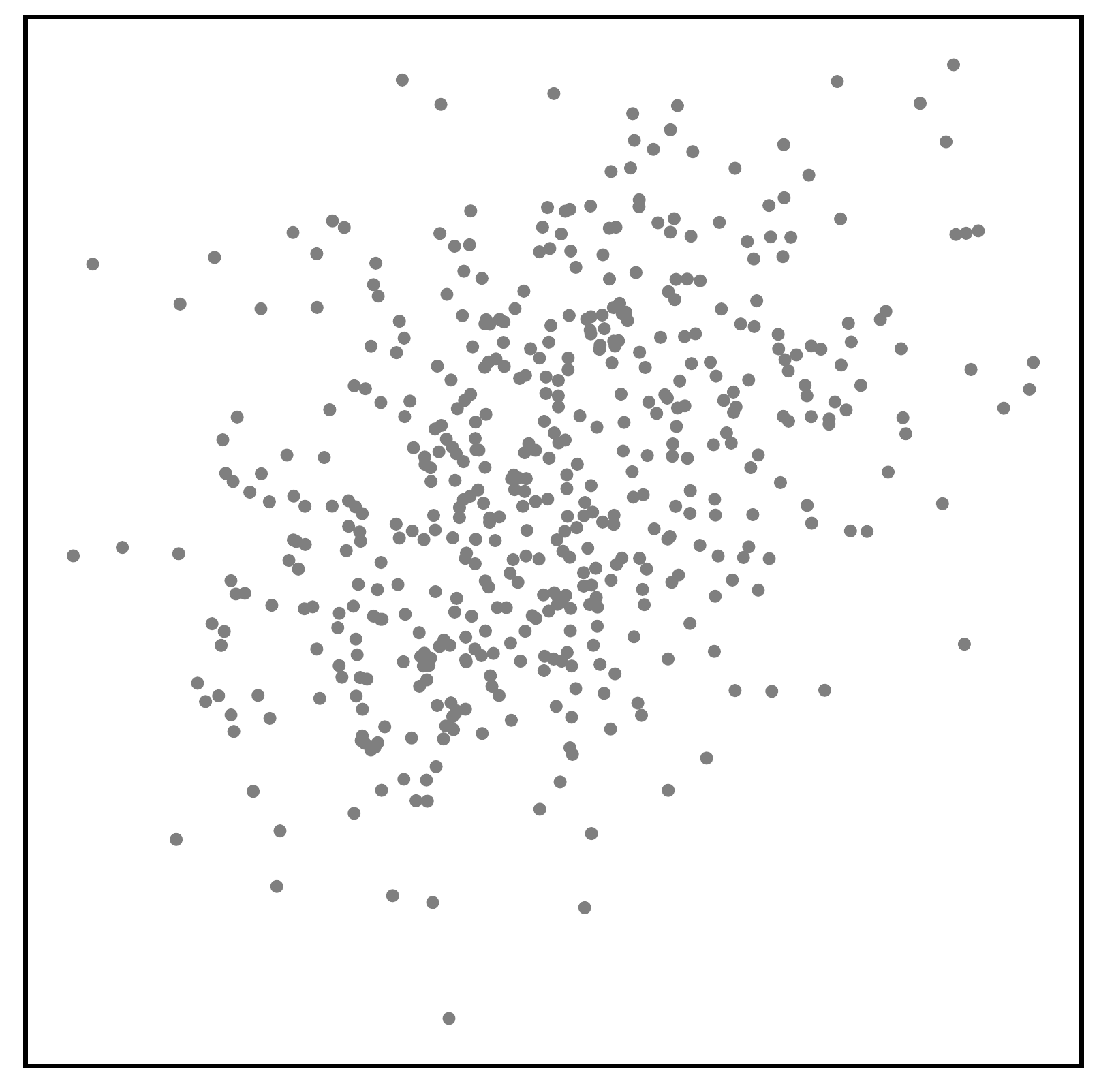}. % 0.23
    \label{tsp:a}
    }
    % \hspace{0.2in}
    \subfigure[]{
    \includegraphics[width=0.12\columnwidth]{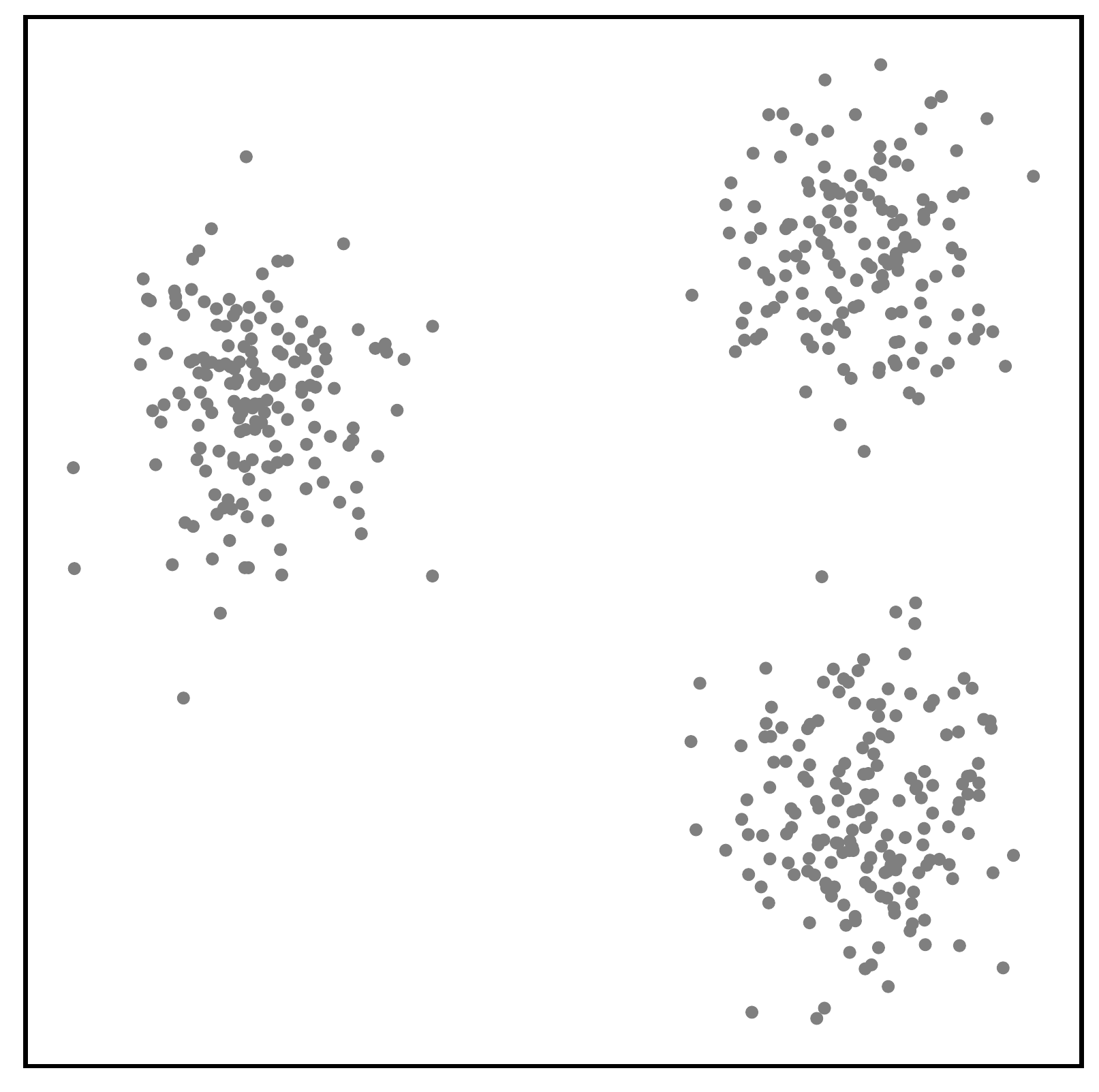}
    \label{tsp:b}
    }
    \subfigure[]{
    \includegraphics[width=0.12\columnwidth]{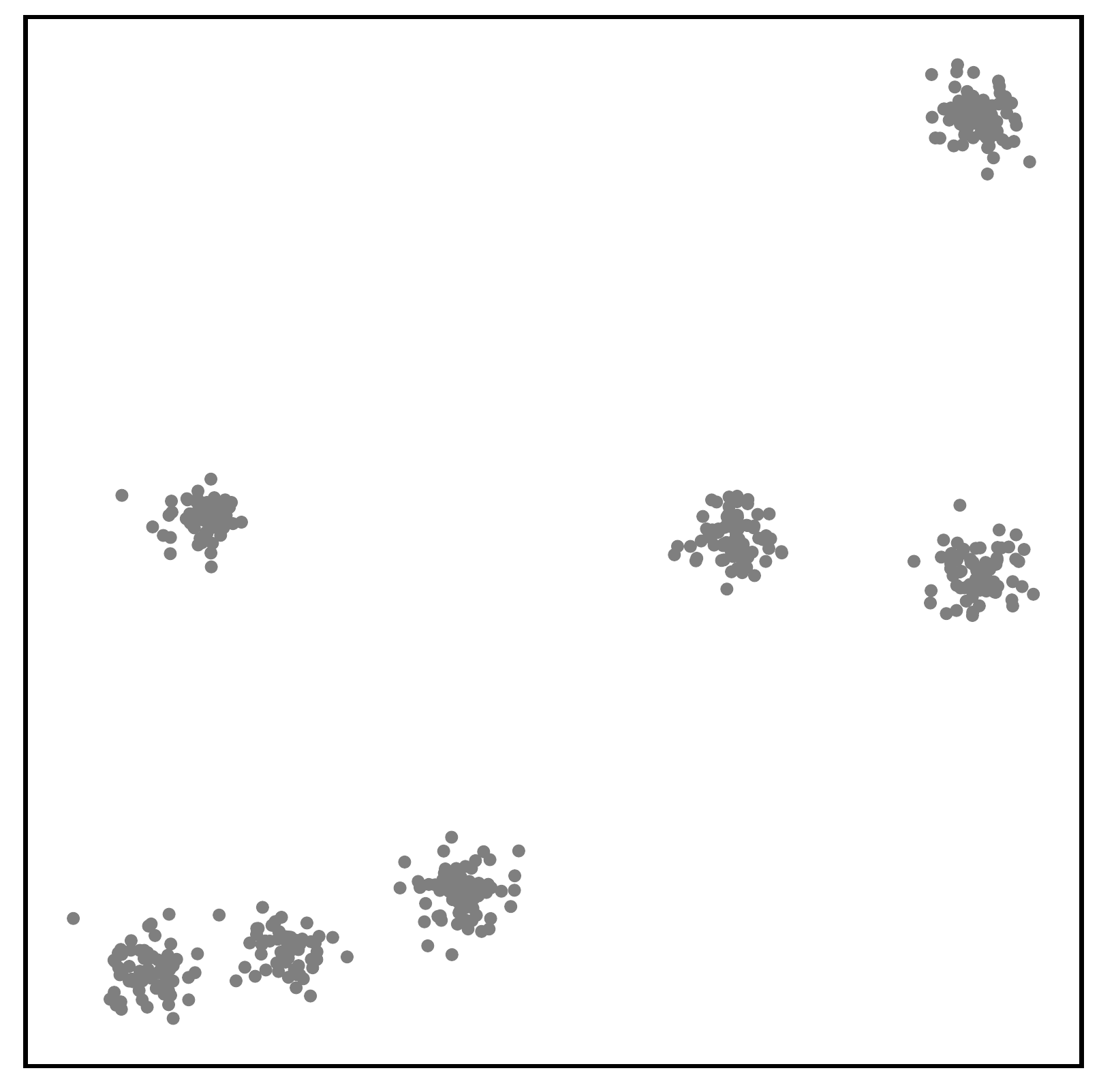}
    \label{tsp:c}
    }\\
    \subfigure[]{
    \includegraphics[width=0.12\columnwidth]{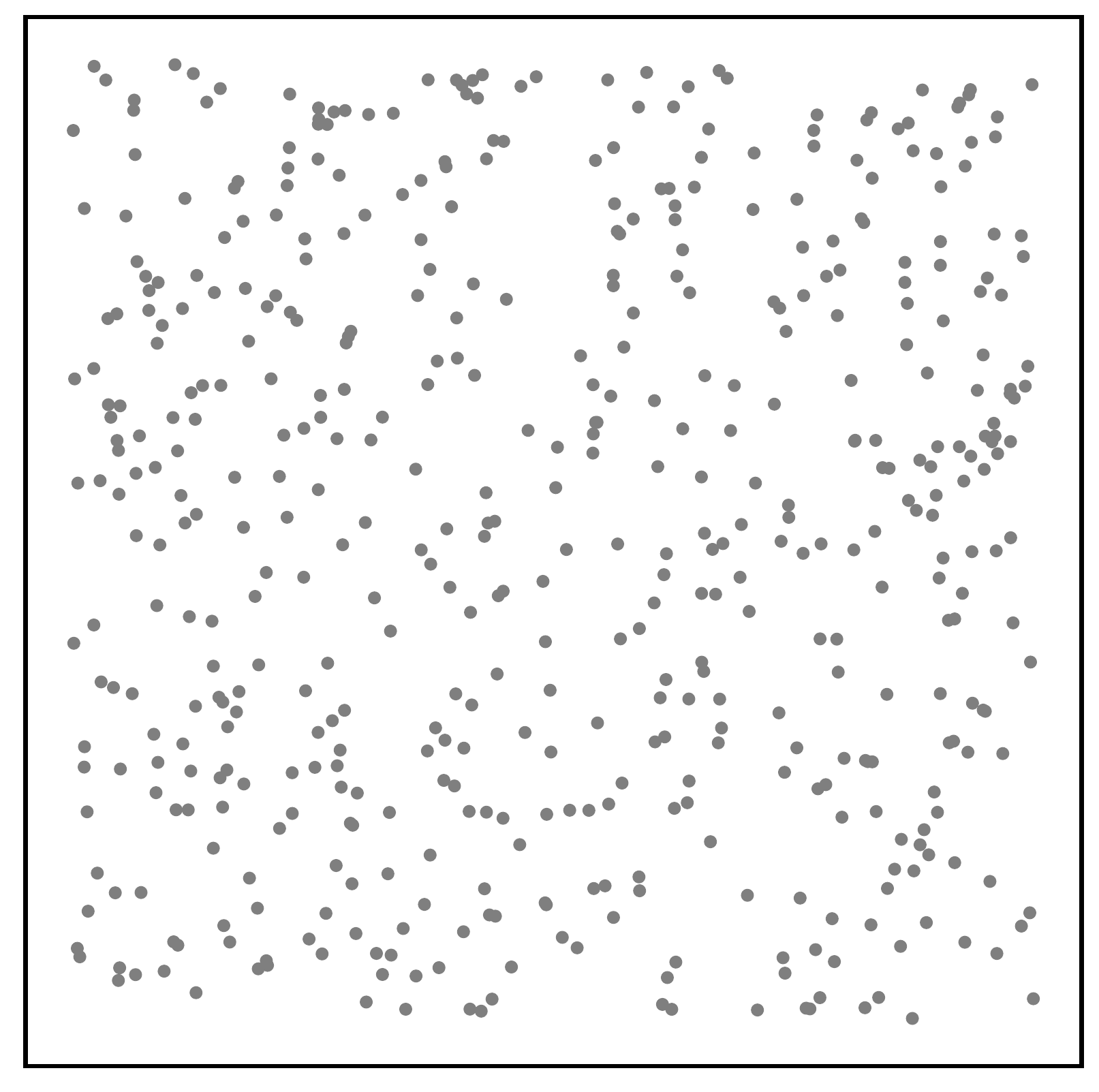}
    \label{tsp:d}
    }
    \subfigure[]{
    \includegraphics[width=0.12\columnwidth]{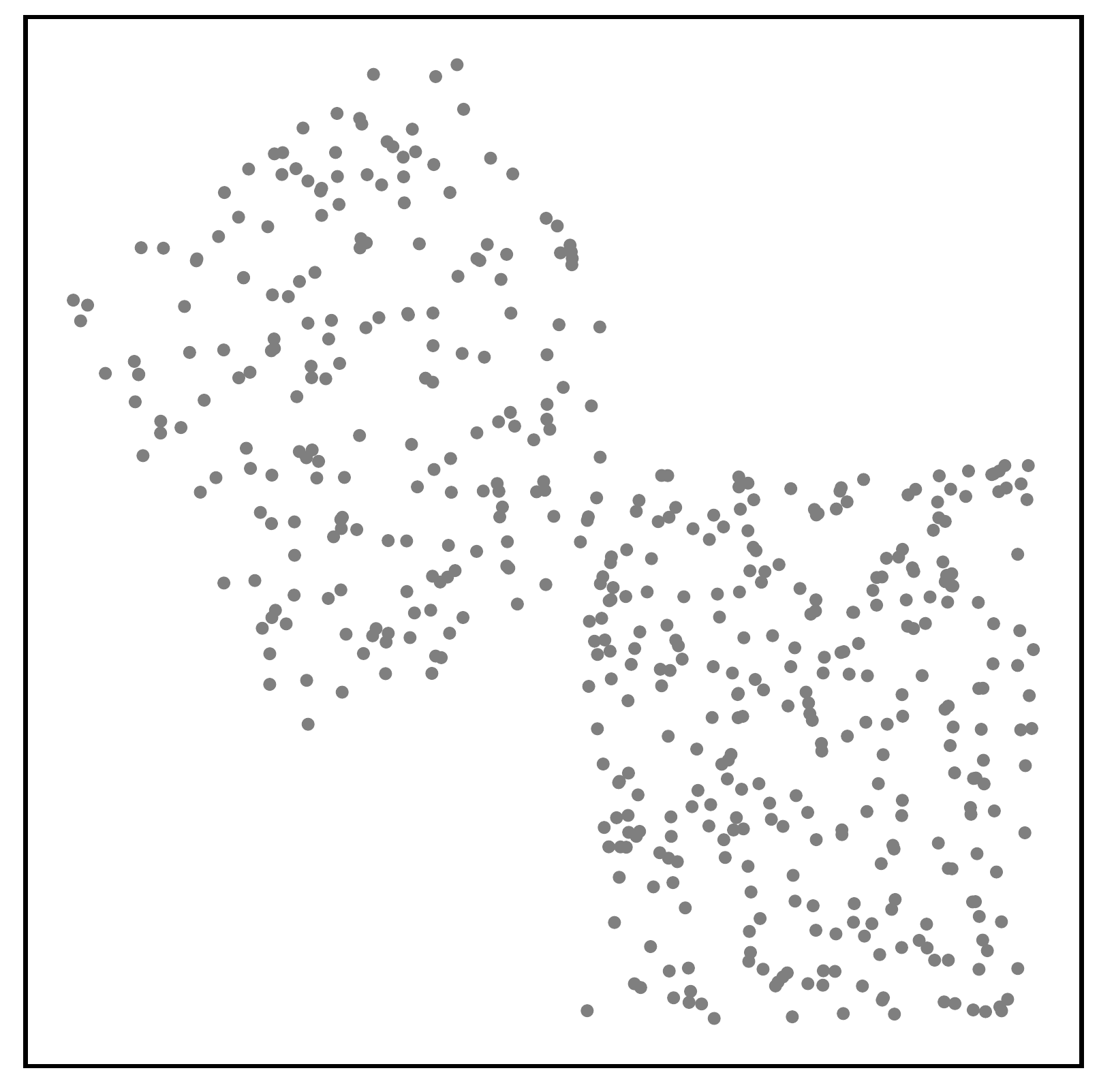}
    \label{tsp:e}
    }
    \subfigure[]{
    \includegraphics[width=0.12\columnwidth]{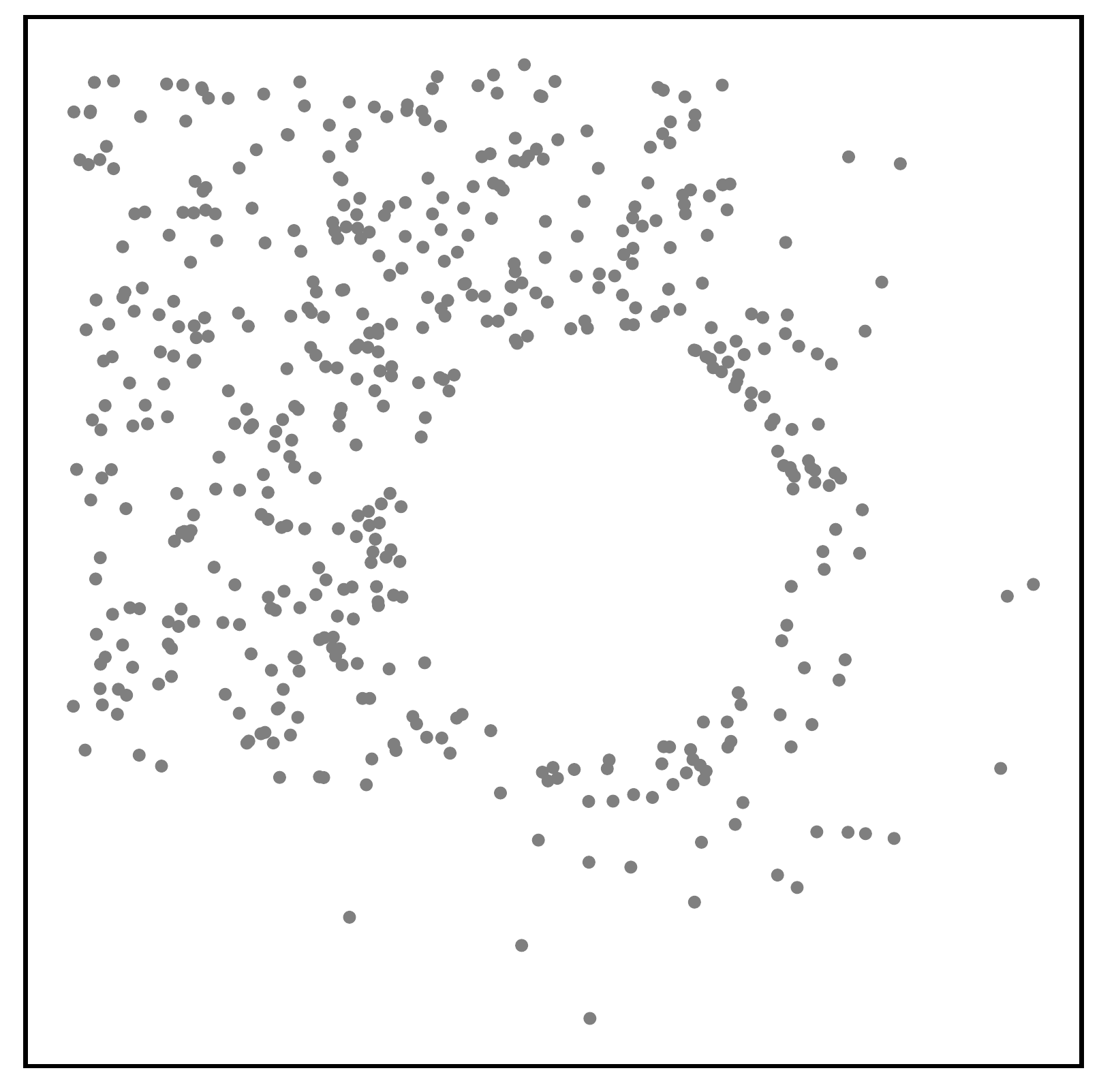}
    \label{tsp:f}
    }
    \vskip 0.05in
    \caption{TSP500 instances following various distributions: (a) $GM_2^5$: Gaussian mixture distribution with $c=2,l=5$; (b) $GM_3^{10}$: Gaussian mixture distribution with $c=3,l=10$; (c) $GM_7^{50}$: Gaussian mixture distribution with $c=7,l=50$; (d) $U$: Uniform distribution; (e) $R$: Rotation distribution; (f) $E$: Explosion distribution.}
    \label{fig_1}
    \end{center}
    \vskip -0.3in
\end{wrapfigure}

We consider VRP instances with various sizes and distributions. Since generating instances with different sizes is relatively easy, here we focus on the details regarding the generation of different distributions of node coordinates. To provide diverse data for meta-training, we generate instances following the uniform distribution, and the gaussian mixture distribution, which is demonstrated to be effective in capturing different hardness levels of the instances~\cite{smith2010understanding}. We further generate instances following rotation and explosion distributions\footnote{\url{https://github.com/jakobbossek/tspgen}} to evaluate the generalization of the learned model. Below, we provide the details on the data generation procedure.

\paragraph{Uniform distribution.} Following the convention~\cite{kool2018attention, kwon2020pomo}, as shown in Figure \ref{tsp:d}, the node coordinate of each node is uniformly sampled from the unit square $U(0, 1)$. 

\paragraph{Gaussian mixture distribution.} Following~\citet{zhang2022learning,manchanda2022generalization}, we parameterize the gaussian mixture distribution with two hyperparameters, cluster $c$ and scale $l$. For the simplicity of notations, we denote it as $GM_c^l$. Specifically, we first generate the coordinate of the center node $v_{c_i}$ of each cluster $c_i$ by uniformly sampling from $U(0,l)$. Other nodes $\mathcal{V} \setminus \{v_{c_i}\}_{i=1}^c$ are then equally distributed into $c$ clusters, where the coordinates of nodes in each cluster forms a (multivariate) gaussian distribution. For example, if a node belongs to the cluster $c_i$, its coordinate is sampled from $\mathcal{N}(n_{c_i}, \textbf{I})$, where the mean is the coordinate of the center node, and the covariance matrix is the identity matrix. Finally, we scale and translate the range of coordinates into $[0,1]$ using the min-max normalization. Some exemplary instances are shown in Figure \ref{tsp:a}-\ref{tsp:c}.

\paragraph{Rotation distribution.} Following~\citet{bossek2019evolving}, we mutate nodes, which originally follow the uniform distribution, by rotating a subset of them (anchored in the origin of the Euclidean plane) as shown in Figure \ref{tsp:e}. The coordinates of selected nodes are transformed by multiplying with $\begin{bmatrix} \cos(\varphi) & \sin(\varphi) \\ -\sin(\varphi) & \cos(\varphi) \end{bmatrix}$, where $\varphi \sim [0, 2\pi]$ is the rotation angle.

\newpage
\paragraph{Explosion distribution.} Following~\citet{bossek2019evolving}, we mutate nodes, which originally follow the uniform distribution, by simulating a random explosion. Specifically, we first randomly select the center of the explosion $v_c$ (i.e., the hole in Figure \ref{tsp:f}). All nodes $v_i$ within the explosion radius $R=0.3$ is moved away from the center with the transformation form of $v_i = v_c + (R+s) \cdot \frac{v_c-v_i}{||v_c-v_i||}$, where $s\sim \text{Exp}(\lambda=1/10)$ is a random value drawn from an exponential distribution.

In this paper, we mainly consider the distribution of node coordinates. For CVRP instances, following~\citet{kool2018attention,kwon2020pomo}, the coordinate of the depot node $v_0$ is uniformly sampled from the unit square $U(0, 1)$. The demand of each node $\delta_i$ is randomly sampled from a discrete uniform distribution $\{1, \cdots, 9\}$. The capacity of each vehicle is set to $Q=\lceil 30+\frac{n}{5} \rceil$, where $n \geq 50$ is the size of CVRP instances. The demand and capacity are further normalized to $\delta_i'=\delta/Q$ and 1, respectively. During meta-training, we generate the diverse task set with hundreds of tasks $\mathcal{T}(n, d)$, where $n\in \mathcal{N}=\{50, 55, \cdots, 200\}$ and $d(c,l) \in \mathcal{D}=\{(0, 0), (1, 1)\} \cup \{c[3,5,7] \times l[10,30,50]\}$. We denote the uniform distribution as $d(c=0,l=0)$. For each size, there are 11 tasks with different distributions in the training task set, and therefore $31 \times 11=341$ tasks in total.

\section{POMO}
\label{app:pomo}
POMO~\cite{kwon2020pomo} significantly improves upon AM~\cite{kool2018attention} by exploiting the symmetry property, which inherently exists in the VRP solution. For example, a solution to a TSP instance is represented as a sequence of nodes. Multiple representations (with different start nodes) exist for the same solution. Previous method~\cite{kool2018attention} selects the start node by the model (using a trainable START token from the NLP community). Since the solution construction process is formulated as a MDP, the first action (i.e., start node) may considerably affect the following actions. However, a desirable model should always be able to construct the optimal solution given different start nodes. POMO considers this symmetry property into the objective function, and the estimated gradient (i.e., Eq. (\ref{eq:reinforce})) could be rewritten as:
\begin{equation}
    \label{eq:pomo_reinforce}
    \small
    \nabla_{\theta} \mathcal{L}(\theta|\mathcal{G}) \approx \frac{1}{S} \sum_{s=1}^{S}(c(\tau_s)-b(\mathcal{G})) \nabla_{\theta} \log p_{\theta}(\tau_s|\mathcal{G}),
\end{equation}
where $\tau_s$ is the solution with start node $v_s \in \mathcal{V}$, $S$ is the number of start nodes (e.g., the size of an instance in TSP), and $b(\mathcal{G})=\frac{1}{S}\sum_{s=1}^S c(\tau_s)$. Intuitively, Eq. (\ref{eq:pomo_reinforce}) forces diverse trajectories towards optimal solution(s). Besides, POMO also leverages instance augmentations to enhance the inference performance. A brief summary is presented in Algorithm \ref{alg:pomo}.
\begin{algorithm}[ht]
  % \footnotesize
  \caption{POMO with REINFORCE}
  \label{alg:pomo}
  {\bf Input:} training instances $\{\mathcal{G}_m\}_{m=1}^M$, number of start nodes $S$, model $\theta$; \\
  {\bf Output:} estimated gradient $\nabla_{\theta}\mathcal{L}(\theta)$;
\begin{algorithmic}[1]
    \STATE $\{v^m_1, v^m_2, \cdots, v^m_S\} \gets$ SelectStartingNodes($\mathcal{G}_m$) \quad $\forall m \in \{1, \cdots, M\}$
    
    \STATE $\tau^m_s \gets \theta(\mathcal{G}_m, v^m_s) \quad \forall m \in \{1, \cdots, M\}, \forall s \in \{1, \cdots, S\}$
    
    \STATE $b(\mathcal{G}_
    m) \gets \frac{1}{S} \sum_{s=1}^S c(\tau^m_s)$ \quad $\forall m \in \{1, \cdots, M\}$ 
    
    % \STATE Obtain $\nabla_{\theta} \mathcal{L}(\theta|\mathcal{G}_m)$ using Eq. (\ref{eq:pomo_reinforce}) \quad $\forall m \in \{1, \cdots, M\}$
    % \STATE $\nabla_{\theta}\mathcal{L}(\theta) \gets \frac{1}{M} \sum_{m=1}^M \nabla_{\theta} \mathcal{L}(\theta|\mathcal{G}_m)$
    
    \STATE $\nabla_{\theta}\mathcal{L}(\theta) \gets \frac{1}{MS} \sum_{m=1}^M \sum_{s=1}^S (c(\tau^m_s)-b(\mathcal{G}_m)) \nabla_{\theta}\log p_{\theta}(\tau^m_s|\mathcal{G}_m)$
\end{algorithmic}
\end{algorithm}

\section{Experiments}
\label{app:exp}
\subsection{Extra Setups}
\label{app:setup}
\paragraph{Setups for Experiments in Figure \ref{story}.} We follow the training setups presented in Section \ref{exp}. For the left panel of Figure \ref{story}, we show the validation result of each method on CVRP (300, $R$). Specifically, \emph{Ours-SO} refers to the second-order method presented in Algorithm \ref{alg:pomo-maml}, \emph{FOMAML} refers to its first-order approximation, where we replace the line 16 of Algorithm \ref{alg:pomo-maml} with Eq. (\ref{eq:fomaml}), and \emph{Ours} refers to the proposed approximation method, where we simply early-stop using the second-order derivative when the training tends to be stable (i.e., at the 50K$_{\mathrm{th}}$ iteration), and leverage the first-order one afterwards.
For the right panel of Figure \ref{story}, the training process follows \emph{Ours-SO}. However, in each iteration of meta-training, besides calculating the direction of the second-order derivative sign$(\nabla_{\theta_0}\mathcal{L}_i(\theta_i^{(K)}))$, we also collect that of FOMAML sign$(\nabla_{\theta_i}\mathcal{L}_i(\theta_i^{(K)}))$ and Reptile sign$(\theta_0-\theta_i^{(K)})$ using the same batch of instances. Note that these two gradients are not used for meta updates. Then, we compute the cosine similarities of these gradient directions with the second-order one (i.e., sign$(\nabla_{\theta_0}\mathcal{L}_i(\theta_i^{(K)}))$), and show the average result over 500 iterations in the right panel of Figure \ref{story}. 

\paragraph{Training Setups for Baselines.} We conduct all experiments on a machine with NVIDIA A100 PCIe (80GB) cards and AMD EPYC 7513 CPU at 2.6GHz. As shown in Section \ref{exp}, we compare our method with several strong baselines. 
% We follow most of their original experimental setups. 
Following the conventional setups in the community~\cite{kool2018attention, kwon2020pomo,hottung2022efficient}, for traditional VRP solvers such as Concorde, LKH3 and HGS, we run them on 32 CPU cores for solving TSP and CVRP instances, while running neural VRP methods on one GPU card. Below, we provide implementation details of all baselines. 
1) Concorde: We use Concorde\footnote{\url{https://www.math.uwaterloo.ca/tsp/concorde/index.html}} Version 03.12.19 with the default setting, to solve TSP instances.
2) LKH3~\cite{helsgaun2017extension}: We use LKH3\footnote{\url{http://akira.ruc.dk/~keld/research/LKH-3/}} Version 3.0.7 to solve TSP and CVRP instances. For each instance, we run LKH3 with 10000 trails and 10 runs. 
3) HGS~\cite{vidal2022hybrid}: We run HGS\footnote{\url{https://github.com/vidalt/HGS-CVRP}} with the default hyperparameters to solve CVRP instances. The maximum number of iterations (without improvement) is set to 20000.
4) For POMO~\cite{kwon2020pomo}, following the training setups presented in Section \ref{exp}, we re-train it for 500K iterations with totally 32M instances, which are randomly sampled from our training task set.
5) AMDKD-POMO~\cite{bi2022learning}
% \footnote{\url{https://github.com/jieyibi/AMDKD}} 
tackles the cross-distribution generalization of neural methods using knowledge distillation. Specifically, it leverages the knowledge from multiple teacher models pretrained on different distributions to yield a generalizable student model. However, it is computationally intractable to obtain a pretrained model for each task since we have hundreds of training tasks on our problem setting. Therefore, following the default setting of~\citet{bi2022learning}, we pretrain three teacher models on instances of size $n=200$, but with distributions chosen from our training task set (i.e., the uniform $U$ and gaussian mixture distributions $GM_3^{10}, GM_7^{50}$). We train each teacher model using 6.4M instances.
% so that the total amount of training instances is similar to other methods. 
After pretraining, we train a light-weight yet generalizable student model by adaptively distilling from the teacher models on another set of 12.8M instances ($n=200$), so that the total amount of training instances (i.e., 32M) is close to other methods. 
% sampled from a subset of our training task set (i.e., $n \in [50, 200]$ and $d \in$ {U, GM_3^{10}, GM_7^{50}}).
6) Meta-POMO~\cite{manchanda2022generalization} leverages Reptile~\cite{nichol2018first}, which does not need to split data into training and validation sets. Therefore, based on its default setting and our experiments in Appendix \ref{app:meta-pomo}, we set $\beta=0.9, B=1, K=50$ and meta-train POMO using Reptile for 10K iterations to keep the same amount of training instances as other methods.
% following the training setups as presented in \ref{exp}, we re-train them with roughly 500K batches of instances (i.e., 32M instances), which are randomly sampled from our training task set.

\subsection{Tuning on Meta-POMO}
\label{app:meta-pomo}
% We find Meta-POMO has a better performance with instance normalization layer.
As shown in Table \ref{sup_exp_1}, we empirically observe the inferior performance of Meta-POMO with a decaying step size $\beta$, which is the straightforward adaptation of~\citet{manchanda2022generalization} to POMO. Specifically, the step size is gradually decayed with the form of $\beta_t = \beta_0 \times \gamma^t$, where $\beta_0=0.99$ is the initial step size, $\gamma=0.999$ is the decay rate and $t$ is the iteration index. 
The undesirable results may be attributed to below factors: a) originally, they only consider around 10 training tasks and randomly select tasks to train, therefore failing to deal with our more complex experimental setting; 
b) further designs may be needed in order to be successfully adapted to POMO, since POMO inherently improves the generalization upon AM. Moreover, the reinforcement learning setting is empirically found to be challenging for Reptile~\cite{nichol2018first}.
We further tune its key hyperparameters (e.g., the step size $\beta$ and number of inner-loop updates $K$) on TSP. We follow the same experimental setups described in Section \ref{exp}, and show the zero-shot performance (i.e., gaps with respect to Concorde) in Table \ref{sup_exp_1}. 
In summary, we empirically observe that a fixed and relatively large step size works better. Therefore, we report the results of Meta-POMO with $\beta=0.9$ and $K=50$ in Section \ref{exp}.
\begin{table*}[!ht]
  \caption{Tuning of Meta-POMO~\cite{manchanda2022generalization} on TSP.}
  \label{sup_exp_1}
  \vskip 0.1in
  \begin{center}
  \begin{small}
%   \begin{sc}
  \renewcommand\arraystretch{1.2}
  \resizebox{\textwidth}{!}{ 
  \begin{tabular}{l|ccccccc|cccc}
    \toprule
    \midrule
    \multirow{2}{*}{Test Task} & \multicolumn{7}{c|}{$K=50$} & \multicolumn{4}{c}{$\beta=0.9$} \\
     & $\beta=0.1$ & $\beta=0.3$ & $\beta=0.5$ & $\beta=0.7$ & $\beta=0.9$ & $\beta=0.99$ & decaying $\beta$ & $K=2$ & $K=5$ & $K=10$ & $K=25$ \\
    \midrule
    % TSP200 (U) & 5.27\% & 3.45\% & 2.84\% & 2.62\% & \textbf{2.25\%} & 0 & 5.48\% & 2.71\% & 2.52\% \\
    $(300, R)$ & 8.53\% & 6.37\% & 5.43\% & 5.03\% & \textbf{4.37\%} & 4.78\% & 9.03\% & 6.18\% & 6.20\% & 5.16\% & 4.89\% \\
    $(300, E)$ & 8.16\% & 5.87\% & 5.05\% & 4.76\% & \textbf{4.14\%} & 4.50\% & 8.50\% & 5.72\% & 5.81\% & 4.78\% & 4.56\% \\
    \midrule
    $(500, R)$ & 14.39\% & 12.10\% & 10.91\% & 10.41\% & \textbf{9.41\%} & 10.12\% & 15.24\% & 11.86\% & 11.91\% & 10.41\% & 10.02\% \\
    $(500, E)$ & 14.16\% & 11.69\% & 10.68\% & 10.46\% & \textbf{9.44\%} & 10.01\% & 14.80\% & 11.55\% & 11.62\% & 10.24\% & 9.94\% \\
    \midrule
    \bottomrule
  \end{tabular}}
%   \end{sc}
  \end{small}
  \end{center}
  \vskip -0.1in
\end{table*}

\subsection{Results on Benchmark Instances}
\label{app:benchmark}
We evaluate all methods on the classical benchmark datasets, such as TSPLIB\footnote{\url{http://comopt.ifi.uni-heidelberg.de/software/TSPLIB95/tsp}} ~\cite{reinelt1991tsplib} and CVRPLIB\footnote{\url{http://vrp.galgos.inf.puc-rio.br/index.php}} (Set-X)~\cite{uchoa2017new}, where we choose representative instances with size $n \in [100, 1002]$. We also combine our method with the efficient active search~\cite{hottung2022efficient}. Specifically, following their original implementation\footnote{\url{https://github.com/ahottung/EAS}}, we run EAS-Lay and EAS-Emb (with 1 run and 200 iterations) on each instance, and report the best result. Due to the huge GPU memory it needs, we only run it on instances with size $n \in [100, 750]$. The detailed results are shown in Table \ref{sup_tsplib} and Table \ref{sup_cvrplib}.

\begin{table*}[!ht]
  \caption{Results on TSPLIB~\cite{reinelt1991tsplib} instances.}
  \label{sup_tsplib}
  \vskip 0.1in
  \begin{center}
  \begin{small}
%   \begin{sc}
  \renewcommand\arraystretch{0.9}
%   \resizebox{\textwidth}{!}{ 
  \begin{tabular}{ll|cccccccc|cc}
    \toprule
    \midrule
    \multicolumn{2}{c|}{} & \multicolumn{2}{c}{POMO} & \multicolumn{2}{c}{AMDKD-POMO} & \multicolumn{2}{c}{Meta-POMO} & \multicolumn{2}{c|}{Ours} & \multicolumn{2}{c}{Ours+EAS}\\
     Instance & Opt. & Obj. & Gap & Obj. & Gap & Obj. & Gap & Obj. & Gap & Obj. & Gap \\
    \midrule
     kroA100 & 21282 & \textbf{21282} & \textbf{0.00\%} & 21360 & 0.37\% & 21308 & 0.12\% & 21305 & 0.11\% & \iffalse21282\fi 21282 & 0.00\% \\
    %  kroB100 & 22141 & 22367 & 1.02\% & 22698 & 2.52\% & 22373 & 1.05\% & 22650 & 2.30\% & 22200 22235 & 0.27\% \\
     kroA150 & 26524 & \textbf{26823} & \textbf{1.13\%} & 26997 & 1.78\% & 26852 & 1.24\% & 26873 & 1.32\% & \iffalse26566\fi 26566 & 0.16\% \\
    %  kroB150 & 26130 & 26337 & 0.79\% & 26704 & 2.20\% & 26282 & 0.58\% & 26452 & 1.23\% & 26147 26147 & 0.07\% \\
     kroA200 & 29368 & \textbf{29745} & \textbf{1.28\%} & 30196 & 2.82\% & 29749 & 1.30\% & 29823 & 1.55\% & \iffalse29481\fi 29460 & 0.31\% \\
     kroB200 & 29437 & 30060 & 2.12\% & 30188 & 2.55\% & 29896 & 1.56\% & \textbf{29814} & \textbf{1.28\%} & 29445 \iffalse29624\fi & 0.03\% \\
     ts225 & 126643 & 131208 & 3.60\% & \textbf{128210} & \textbf{1.24\%} & 131877 & 4.13\% & 128770 & 1.68\% & 127281 \iffalse127396\fi & 0.50\% \\
     tsp225 & 3916 & 4040 & 3.17\% & 4074 & 4.03\% & 4047 & 3.35\% & \textbf{4008} & \textbf{2.35\%} & 3933 \iffalse3946\fi & 0.43\% \\
     pr226 & 80369 & \textbf{81509} & \textbf{1.42\%} & 82430 & 2.56\% & 81968 & 1.99\% & 81839 & 1.83\% & 81235 \iffalse81397\fi & 1.08\% \\
     pr264 & 49135 & 50513 & 2.80\% & 51656 & 5.13\% & \textbf{50065} & \textbf{1.89\%} & 50649 & 3.08\% & 49212 \iffalse49666\fi & 0.16\% \\
     a280 & 2579 & 2714 & 5.23\% & 2773 & 7.52\% & 2703 & 4.81\% & \textbf{2695} & \textbf{4.50\%} & 2591 \iffalse2599\fi & 0.47\% \\
     pr299 & 48191 & 50571 & 4.94\% & 51270 & 6.39\% & 49773 & 3.28\% & \textbf{49348} & \textbf{2.40\%} & 48449 \iffalse48511\fi & 0.54\% \\
     lin318 & 42029 & 44011 & 4.72\% & 44154 & 5.06\% & \textbf{43807} & \textbf{4.23\%} & 43828 & 4.28\% & \iffalse43097\fi 43090 & 2.52\% \\
     rd400 & 15281 & 16254 & 6.37\% & 16610 & 8.70\% & 16153 & 5.71\% & \textbf{15948} & \textbf{4.36\%} & \iffalse15823\fi 15531 & 1.64\% \\
     fl417 & 11861 & 12940 & 9.10\% & 13129 & 10.69\% & 12849 & 8.33\% & \textbf{12683} & \textbf{6.93\%} & 12754 \iffalse13012\fi & 7.53\% \\
     pr439 & 107217 & 115651 & 7.87\% & 117872 & 9.94\% & 114872 & 7.14\% & \textbf{114487} & \textbf{6.78\%} & 111902 \iffalse112186\fi & 4.37\% \\
     pcb442 & 50778 & 55273 & 8.85\% & 56225 & 10.73\% & 55507 & 9.31\% & \textbf{54531} & \textbf{7.39\%} & \iffalse53425\fi 53069 & 4.51\% \\
     d493 & 35002 & 38388 & 9.67\% & 38400 & 9.71\% & 38641 & 10.40\% & \textbf{38169} & \textbf{9.05\%} & \iffalse37874\fi 37850 & 8.14\% \\
     u574 & 36905 & 41574 & 12.65\% & 41426 & 12.25\% & 41418 & 12.23\% & \textbf{40515} & \textbf{9.78\%} & \iffalse40348\fi 39295 & 6.48\% \\
     rat575 & 6773 & \textbf{7617} & \textbf{12.46\%} & 7707 & 13.79\% & 7620 & 12.51\% & 7658 & 13.07\% & \iffalse7627\fi 7333 & 8.27\% \\
     p654 & 34643 & 38556 & 11.30\% & 39327 & 13.52\% & 38307 & 10.58\% & \textbf{37488} & \textbf{8.21\%} & 39141 \iffalse41391\fi & 12.98\% \\
     d657 & 48912 & 55133 & 12.72\% & 55143 & 12.74\% & 54715 & 11.86\% & \textbf{54346} & \textbf{11.11\%} & \iffalse54403\fi 53077 & 8.52\% \\
     u724 & 41910 & 48855 & 16.57\% & 48738 & 16.29\% & 48272 & 15.18\% & \textbf{48026} & \textbf{14.59\%} & \iffalse48976\fi 48144 & 14.87\% \\
     rat783 & 8806 & 10401 & 18.11\% & 10338 & 17.40\% & \textbf{10228} & \textbf{16.15\%} & 10300 & 16.97\% & \multicolumn{2}{c}{--} \iffalse10590 10428 20.26\%\fi \\
     pr1002 & 259045 & 310855 & 20.00\% & 312299 & 20.56\% & 308281 & 19.01\% & \textbf{305777} & \textbf{18.04\%} & \multicolumn{2}{c}{--} \iffalse318045 323239 22.78\%\fi \\
    % \midrule
    %  \multicolumn{2}{c|}{Avg. Gap} & \multicolumn{2}{c}{7.66\%} & \multicolumn{2}{c}{0} & \multicolumn{2}{c}{7.23\%} & \multicolumn{2}{c|}{6.55\%} & \multicolumn{2}{c}{--} \\
    \midrule
    \bottomrule
  \end{tabular}
%   \end{sc}
  \end{small}
  \end{center}
  \vskip -0.15in
\end{table*}

\begin{table*}[!ht]
  \caption{Results on CVRPLIB (Set-X)~\cite{uchoa2017new} instances.}
  \label{sup_cvrplib}
  \vskip 0.1in
  \begin{center}
  \begin{small}
%   \begin{sc}
  \renewcommand\arraystretch{0.9}
%   \resizebox{\textwidth}{!}{ 
  \begin{tabular}{ll|cccccccc|cc}
    \toprule
    \midrule
    \multicolumn{2}{c|}{} & \multicolumn{2}{c}{POMO} & \multicolumn{2}{c}{AMDKD-POMO} & \multicolumn{2}{c}{Meta-POMO} & \multicolumn{2}{c|}{Ours} & \multicolumn{2}{c}{Ours+EAS}\\
     Instance & Opt. & Obj. & Gap & Obj. & Gap & Obj. & Gap & Obj. & Gap & Obj. & Gap \\
    \midrule
     X-n101-k25 & 27591 & \textbf{28804} & \textbf{4.40\%} & 28947 & 4.91\% & 29647 & 7.45\% & 29442 & 6.71\% & 27750 \iffalse27979\fi & 0.58\% \\
    %  X-n148-k46 & 43448 & 45672 & 5.12\% & 45617 & 0 & 46540 & 0\% & 46438 & 6.88\% & 43943 & 0 \\
     X-n153-k22 & 21220 & 23701 & 11.69\% & 23179 & 9.23\% & 23428 & 10.41\% & \textbf{22810} & \textbf{7.49\%} & 21864 \iffalse21942\fi & 3.03\% \\
     X-n200-k36 & 58578 & \textbf{60983} & \textbf{4.11\%} & 61074 & 4.26\% & 61632 & 5.21\% & 61496 & 4.98\% & \iffalse60040\fi 59765 & 2.03\% \\
     X-n251-k28 & 38684 & \textbf{40027} & \textbf{3.47\%} & 40262 & 4.08\% & 40477 & 4.63\% & 40059 & 3.55\% & 39198 \iffalse39231\fi & 1.33\% \\
     X-n303-k21 & 21736 & 22724 & 4.55\% & 22861 & 5.18\% & 22661 & 4.26\% & \textbf{22624} & \textbf{4.09\%} & \iffalse22114\fi 22035 & 1.38\% \\
     X-n351-k40 & 25896 & \textbf{27410} & \textbf{5.85\%} & 27431 & 5.93\% & 27992 & 8.09\% & 27515 & 6.25\% & \iffalse26654\fi 26644 & 2.89\% \\
     X-n401-k29 & 66154 & 68435 & 3.45\% & 68579 & 3.67\% & 68272 & 3.20\% & \textbf{68234} & \textbf{3.14\%} & \iffalse67507\fi 67365 & 1.83\% \\
     X-n459-k26 & 24139 & 26612 & 10.24\% & 26255 & 8.77\% & 25789 & 6.84\% & \textbf{25706} & \textbf{6.49\%} & \iffalse25151\fi 25144 & 4.16\% \\
     X-n502-k39 & 69226 & 71435 & 3.19\% & 71390 & 3.13\% & 71209 & 2.86\% & \textbf{70769} & \textbf{2.23\%} & 70277 \iffalse70616\fi & 1.52\% \\
     X-n548-k50 & 86700 & 90904 & 4.85\% & 90890 & 4.83\% & 90743 & 4.66\% & \textbf{90592} & \textbf{4.49\%} & \iffalse89644\fi 89542 & 3.28\% \\
     X-n599-k92 & 108451 & 115894 & 6.86\% & 115702 & 6.69\% & \textbf{115627} & \textbf{6.62\%} & 116964 & 7.85\% & 113089 \iffalse113612\fi & 4.28\% \\
     X-n655-k131 & 106780 & 110327 & 3.32\% & 111587 & 4.50\% & 110756 & 3.72\% & \textbf{110096} & \textbf{3.11\%} & 108433 \iffalse109088\fi & 1.55\% \\
     X-n701-k44 & 81923 & 86933 & 6.12\% & 88166 & 7.62\% & 86605 & 5.72\% & \textbf{86005} & \textbf{4.98\%} & 85432 \iffalse86145\fi & 4.28\% \\
     X-n749-k98 & 77269 & \textbf{83294} & \textbf{7.80\%} & 83934 & 8.63\% & 84406 & 9.24\% & 83893 & 8.57\% & 81040 \iffalse81798\fi & 4.88\% \\
     X-n801-k40 & 73311 & 80584 & 9.92\% & 80897 & 10.35\% & 79077 & 7.87\% & \textbf{78171} & \textbf{6.63\%} & \multicolumn{2}{c}{--} \iffalse78474 0\fi \\
     X-n856-k95 & 88965 & 96398 & 8.35\% & 95809 & 7.69\% & \textbf{95801} & \textbf{7.68\%} & 96739 & 8.74\% & \multicolumn{2}{c}{--} \iffalse93466 0\fi \\
     X-n895-k37 & 53860 & 61604 & 14.38\% & 62316 & 15.70\% & 59778 & 10.99\% & \textbf{58947} & \textbf{9.44\%} & \multicolumn{2}{c}{--} \iffalse59081 0\fi \\
    %  X-n916-k207 & 329179 & 352793 & 7.17\% & 361058 & 0 & 354526 & 7.70\% & 352581 & 7.11\% & \multicolumn{2}{c}{--} \\
    %  X-n936-k151 & 132715 & 160235 & 20.74\% & 156021 & 0 & 156530 & 17.94\% & 157898 & 18.98\% & \multicolumn{2}{c}{--} \\
     X-n957-k87 & 85465 & 93221 & 9.08\% & 93995 & 9.98\% & 92647 & 8.40\% & \textbf{92011} & \textbf{7.66\%} & \multicolumn{2}{c}{--} \\
    %  X-n979-k58 & 118976 & 127555 & 7.21\% & 127038 & 0 & 127148 & 6.87\% & 125289 & 5.31\% & \multicolumn{2}{c}{--} \\
     X-n1001-k43 & 72355 & 82046 & 13.39\% & 82855 & 14.51\% & 79347 & 9.66\% & \textbf{78955} & \textbf{9.12\%} & \multicolumn{2}{c}{--} \\
    % \midrule
    %  \multicolumn{2}{c|}{Avg. Gap} & \multicolumn{2}{c}{7.11\%} & \multicolumn{2}{c}{0} & \multicolumn{2}{c}{7.41\%} & \multicolumn{2}{c|}{6.08\%} & \multicolumn{2}{c}{0} \\
    \midrule
    \bottomrule
  \end{tabular}
%   \end{sc}
  \end{small}
  \end{center}
  \vskip -0.1in
\end{table*}

\newpage
We further evaluate all methods on Set-XML100~\cite{queiroga202210}, which is a newly proposed CVRP benchmark dataset, with the size of instances $n \in [100, 5000]$. The instances have a broader range of distribution shifts, such as depot positioning ($A$), customer positioning ($B$), demand distribution ($C$), and average route size ($D$). Since the original dataset only contains VRP100 instances, we randomly sample 5 instances and further generate 30 ($=5\times 6$) instances with size $n \in \{500, 1000, 2000, 3000, 4000, 5000\}$. We randomly sample the four characteristics from the Cartesian product of $A\in \{1,2,3\} \times B\in \{1,2,3\} \times C\in \{1,2,3,4,5,6,7\} \times D\in \{1,2,3,4,5,6\}$. We use the given optimal solutions for CVRP100 instances, and use HGS to obtain (sub-)optimal solutions for other newly generated instances. The setting of HGS is the same as the one in Appendix D.1. The solving time varies from hours to days depends on $n$. The results are shown in Table \ref{sup_cvrplib_xml100}, where each instance has the form of $\text{XML}\{n\}\_\{ABCD\}\_\{\text{ID}\}$. We omit the ID (i.e., 01) for simplicity.

\begin{table*}[!ht]
  \caption{Results on CVRPLIB (Set-XML100)~\cite{queiroga202210} instances.}
  \label{sup_cvrplib_xml100}
  \vskip 0.1in
  \begin{center}
  \begin{small}
%   \begin{sc}
  \renewcommand\arraystretch{1.25}
%   \resizebox{\textwidth}{!}{ 
  \begin{tabular}{ll|cccccccc}
    \toprule
    \midrule
    \multicolumn{2}{c|}{} & \multicolumn{2}{c}{POMO} & \multicolumn{2}{c}{AMDKD-POMO} & \multicolumn{2}{c}{Meta-POMO} & \multicolumn{2}{c}{Ours}\\
     Instance  &  (Sub-)Opt. & Obj. & Gap & Obj. & Gap & Obj. & Gap & Obj. & Gap \\
    \midrule
     XML100\_1113 & 14740 & \textbf{15049} & \textbf{2.10\%} & 15182 & 3.00\% & 15125 & 2.61\% & 15076 & 2.28\% \\
     XML100\_1341 & 24931 & 25927 & 4.00\% & \textbf{25796} & \textbf{3.47\%} & 26560 & 6.53\% & 26143 & 4.86\% \\
     XML100\_2271 & 20100 & 21782 & 8.37\% & 21109 & 5.02\% & 21333 & 6.13\% & \textbf{20877} & \textbf{3.87\%} \\
     XML100\_3123 & 20370 & \textbf{20704} & \textbf{1.64\%} & 20978 & 2.98\% & 20907 & 2.64\% & 20883 & 2.52\% \\
     XML100\_3372 & 33926 & 37235 & 9.75\% & 37301 & 9.95\% & 37082 & 9.30\% & \textbf{36292} & \textbf{6.97\%} \\
     XML500\_1215 & 37174 & 39302 & 5.72\% & 39152 & 5.32\% & 38817 & 4.42\% & \textbf{38689} & \textbf{4.08\%} \\
     XML500\_1246 & 23205 & 25532 & 10.03\% & 25516 & 9.96\% & 25212 & 8.65\% & \textbf{25096} & \textbf{8.15\%} \\
     XML500\_1344 & 47944 & 51257 & 6.91\% & 51452 & 7.32\% & \textbf{50541} & \textbf{5.42\%} & 50657 & 5.66\% \\
     XML500\_3134 & 65408 & 69527 & 6.30\% & 69675 & 6.52\% & 69284 & 5.93\% & \textbf{68703} & \textbf{5.04\%} \\
     XML500\_3315 & 44783 & 47556 & 6.19\% & 47595 & 6.28\% & 47294 & 5.61\% & \textbf{47104} & \textbf{5.18\%} \\
     XML1000\_1276 & 42095 & 48226 & 14.56\% & 49132 & 16.72\% & 46358 & 10.13\% & \textbf{46342} & \textbf{10.09\%} \\
     XML1000\_1335 & 63968 & 72555 & 13.42\% & 72733 & 13.70\% & 70118 & 9.61\% & \textbf{69470} & \textbf{8.60\%} \\
     XML1000\_2256 & 30862 & 36202 & 17.30\% & 36448 & 18.10\% & 34908 & 13.11\% & \textbf{34182} & \textbf{10.76\%} \\
     XML1000\_2363 & 85618 & 96685 & 12.93\% & 95985 & 12.11\% & 94893 & 10.83\% & \textbf{93445} & \textbf{9.14\%} \\
     XML1000\_3113 & 169377 & 179276 & 5.84\% & 180583 & 6.62\% & 178765 & 5.54\% & \textbf{178171} & \textbf{5.19\%} \\
     XML2000\_1172 & 336322 & \textbf{392613} & \textbf{16.74\%} & 416007 & 23.69\% & 414319 & 23.19\% & 395090 & 17.47\% \\
     XML2000\_1214 & 194617 & 209676 & 7.74\% & 211107 & 8.47\% & 206678 & 6.20\% & \textbf{205204} & \textbf{5.44\%} \\
     XML2000\_1326 & 69613 & 97656 & 40.28\% & 95535 & 37.24\% & 84193 & 20.94\% & \textbf{83356} & \textbf{19.74\%} \\
     XML2000\_2216 & 56550 & 75417 & 33.36\% & 70690 & 25.00\% & 65542 & 15.90\% & \textbf{63906} & \textbf{13.01\%} \\
     XML2000\_3316 & 105108 & 120956 & 15.08\% & 129375 & 23.09\% & 119440 & 13.64\% & \textbf{116758} & \textbf{11.08\%} \\
     XML3000\_1141 & 800995 & \textbf{890313} & \textbf{11.15\%} & 980642 & 22.43\% & 938765 & 17.20\% & 910961 & 13.73\% \\
     XML3000\_2221 & 615170 & 667875 & 8.57\% & 703465 & 14.35\% & \textbf{656973} & \textbf{6.80\%} & 674764 & 9.69\% \\
     XML3000\_2322 & 400934 & 450847 & 12.45\% & 487873 & 21.68\% & 448146 & 11.78\% & \textbf{446922} & \textbf{11.47\%} \\
     XML3000\_3155 & 244524 & 328102 & 34.18\% & 308877 & 26.32\% & 285693 & 16.84\% & \textbf{271352} & \textbf{10.97\%} \\
     XML3000\_3313 & 427510 & 471327 & 10.25\% & 488874 & 14.35\% & 467088 & 9.26\% & \textbf{459396} & \textbf{7.46\%} \\
     XML4000\_1211 & 1296150 & 1397205 & 7.80\% & 1451127 & 11.96\% & 1360158 & 4.94\% & \textbf{1336333} & \textbf{3.10\%} \\
     XML4000\_1246 & 149850 & 190303 & 27.00\% & 198247 & 32.30\% & \textbf{173269} & \textbf{15.63\%} & 174495 & 16.45\% \\
     XML4000\_2153 & 330364 & 684832 & 107.30\% & 540420 & 63.58\% & \textbf{379186} & \textbf{14.78\%} & 502960 & 52.24\% \\
     XML4000\_3161 & 1516100 & 1694469 & 11.76\% & 1805507 & 19.09\% & 1755874 & 15.82\% & \textbf{1658308} & \textbf{9.38\%} \\
     XML4000\_3246 & 156226 & 292968 & 87.53\% & 206601 & 32.24\% & 184648 & 18.19\% & \textbf{183801} & \textbf{17.65\%} \\
     XML5000\_1241 & 1584020 & 1741474 & 9.94\% & 1778777 & 12.30\% & 1826274 & 15.29\% & \textbf{1718791} & \textbf{8.51\%} \\
     XML5000\_1321 & 1466910 & \textbf{1597897} & \textbf{8.93\%} & 1886757 & 28.62\% & 1677306 & 14.34\% & 1647989 & 12.34\% \\
     XML5000\_2224 & 315739 & 386773 & 22.50\% & 403097 & 27.67\% & \textbf{345626} & \textbf{9.47\%} & 352382 & 11.61\% \\
     XML5000\_3135 & 396487 & 892760 & 125.17\% & 755735 & 90.61\% & 477590 & 20.46\% & \textbf{449778} & \textbf{13.44\%} \\
     XML5000\_3372 & 1135140 & \textbf{1273886} & \textbf{12.22\%} & 1522414 & 34.12\% & 1438497 & 26.72\% & 1293027 & 13.91\% \\
    \midrule
    \bottomrule
  \end{tabular}
%   \end{sc}
  \end{small}
  \end{center}
  \vskip -0.1in
\end{table*}

\newpage
\subsection{Ablation Study}
\label{app:ablation}
Here, we conduct further ablation studies on hyperparameters and technical choices. For training efficiency, we run the experiments for 125K iterations (i.e., 16M instances) using the proposed first-order approximation method on TSP. The other training setups are kept the same as the ones presented in Section \ref{exp}.

\textbf{Ablation Study on Hyperparameters.}
There are several key hyperparameters in the proposed framework: 1) the number of tasks in a mini-batch $B$; 2) the number of inner-loop updates $K$; 3) the step sizes of inner-loop and outer-loop updates $\alpha, \beta$; 4) the temperature $\eta$, which controls the entropy of the probability distribution, from which our hierarchical task scheduler samples. The results are shown in Table \ref{app_exp_abl0}. 
Note that the performance comparison is based on the same number of training instances. Since the training instances in the inner-loop and outer-loop optimization are different, increasing $B$ or $K$ will decrease the total number of meta-model updates in the outer-loop optimization, resulting in inferior zero-shot performance. In this paper, we follow the setting in~\citet{finn2017model}, with $B=K=1$. If we would like to increase $B$ or $K$, 1) the step sizes $\alpha, \beta$ need to be tuned in order to achieve decent performance; 2) the advanced meta-learning algorithms (e.g., MAML with bootstrapping~\cite{flennerhag2022bootstrapped}) could be used to improve training efficiency. Moreover, we could observe that carefully tuning the temperature $\eta$ may further boost the performance.

\begin{table}[!ht]
  \vskip -0.15in
  \caption{Ablation study on Hyperparameters.}
  \label{app_exp_abl0}
  \vskip 0.1in
  \begin{center}
  \begin{small}
  % \resizebox{0.45\textwidth}{!}{ 
  \begin{tabular}{ccccc|cc}
    \toprule
    \midrule
    $B$ & $K$ & $\alpha$ & $\beta$ & $\eta$ & $(300, U)$ & $(500, R)$ \\
    \midrule
    1 & 1 & 1e-4 & 1e-4 & 1 & 13.51 (4.26\%) & 13.54 (9.29\%) \\
    1 & 1 & 1e-4 & 1e-3 & 1 & 13.61 (5.03\%) & 13.71 (10.61\%) \\
    1 & 3 & 1e-4 & 1e-4 & 1 & 14.04 (8.41\%) & 13.45 (14.61\%) \\
    3 & 1 & 1e-4 & 1e-4 & 1 & 14.00 (8.09\%) & 14.39 (16.16\%) \\
    1 & 1 & 1e-4 & 1e-4 & 0.2 & \textbf{13.49 (4.13\%)} & \textbf{13.53 (9.18\%)} \\
    1 & 1 & 1e-4 & 1e-4 & 5 & 13.62 (5.07\%) & 13.73 (10.80\%) \\
    \midrule
    \bottomrule
  \end{tabular}
%   \end{sc}
  \end{small}
  \end{center}
  \vskip -0.1in
\end{table}

\textbf{Ablation Study on Optimizers.}
1) \emph{Optimizer:} when training with REINFORCE~\cite{williams1992simple}, the Adam optimizer has much better performance than SGD, as shown in~\citet{kool2018attention,kwon2020pomo}. We also empirically observe the superior performance when meta-training POMO with the Adam optimizer in the out-loop optimization (where the meta-model is updated). Another question is \emph{which optimizer should we use in the inner-loop optimization?} We conduct the ablation study pertaining to this question, and the results in Table \ref{app_exp_abl1} demonstrate similar performance for different optimizers in the inner loop. 
Therefore, in this paper, we use the same optimizer (i.e., Adam) as the one used during fine-tuning, which is more convenient than tuning different optimizers in the same framework.
2) \emph{Meta-Gradient:} the Adam optimizer~\cite{kingma2015adam} is known for its performance and stability, and requires fewer hyperparameters for tuning. The internal implementation of Adam incorporates bias correction, scaling and momentum. Here the question is \emph{could it be helpful to load the information of meta-gradients to the inner-loop optimizer?} The intuition is that the information of the meta-gradient (from the outer-loop optimizer) may serve as a good initialization of the gradient for the inner-loop optimization, so that the meta-training may achieve better convergence or final performance. Based on the results in Table \ref{app_exp_abl1}, we observe that loading meta-gradients has no improvements. Therefore we ignore it to keep the simplicity of the framework.

\begin{table}[!ht]
  \vskip -0.15in
  \caption{Ablation study on Optimizers.}
  \label{app_exp_abl1}
  \vskip 0.1in
  \begin{center}
  \begin{small}
  % \resizebox{0.45\textwidth}{!}{ 
  \begin{tabular}{ccc|cc}
    \toprule
    \midrule
    Outer-Loop & Inner-Loop & Load Meta-Gradient & $(300, U)$ & $(500, R)$ \\
    \midrule
    Adam & Adam & $\times$ & \textbf{13.51 (4.26\%)} & \textbf{13.54 (9.29\%)} \\
    SGD & SGD & $\times$ & 16.56 (27.83\%) & 17.47 (41.16\%) \\
    Adam & SGD & $\times$ & 13.51 (4.32\%) & 13.55 (9.37\%) \\
    Adam & Adam & $\checkmark$ & 13.53 (4.47\%) & 13.60 (9.72\%) \\
    \midrule
    \bottomrule
  \end{tabular}
%   \end{sc}
  \end{small}
  \end{center}
  \vskip -0.1in
\end{table}

\textbf{Ablation Study on Normalization Layers.}
Similar to~\citet{drakulic2023bq}, we empirically observe that the choice of the normalization layer (in attention-based models~\cite{kool2018attention,kwon2020pomo}) has a significant effect on the final performance. We denote the batch normalization without tracking the running mean and variance as \emph{batch\_no\_track}. 
Empirically, the batch\_no\_track and instance normalization could achieve decent zero-shot performance, while no normalization or batch normalization (with the first-order approximation) may destabilize the meta-training. We also try the rezero normalization layer~\cite{bachlechner2021rezero} without observing significant improvements. Therefore, in this paper, we use batch\_no\_track as the default normalization layer for our method. 
Although the above empirical observations may only valid for attention-based models, the choice of the normalization layer may be worthy of attention for the future work.

\subsection{Generalizability}
\label{app:generalizability}
To demonstrate the generalizability of the proposed framework, we further apply it to L2D\footnote{\url{https://github.com/mit-wu-lab/learning-to-delegate}}~\cite{li2021learning}.
Concretely, we train a regression model (rather than a classification model) since: (1) the training of the regression model is more efficient (i.e., around 6 hrs); (2) it has better flexibility in training multiple sizes and distributions, which is quite suitable for the omni-generalization setting. We use the datasets provided by~\citet{li2021learning} to construct the training task set. Concretely, it contains six mixed CVRP distributions with $n\in \{500, 1000\} \times d \in \{3,5,7\}$, where $n$ is the problem size and $d$ is the cluster center. We use HGS as the subsolver, and keep the other settings the same as~\citet{li2021learning}. During the evaluation, we set the number of runs to 1 for each instance, and set the time limit for solving each subproblem to 1s.
For a fair comparison, we retrain the regression model (i.e., L2D with the batch size of 512 and 2048), and meta-train a regression model (i.e., Ours with the batch size of 512) on the training task set. 
We show the average cost over 10 instances on each test dataset, and the number of achieved best solutions (in brackets) among all methods.
As shown in Table \ref{app_exp_gene}, our method could further improve the generalization of L2D when meta-training with diverse tasks in terms of sizes and distributions, demonstrating the effectiveness and generalizability of the proposed framework.

\begin{table}[!ht]
  \vskip -0.1in
  \caption{Performance evaluation on L2D~\cite{li2021learning}.}
  \label{app_exp_gene}
  \vskip 0.1in
  \begin{center}
  % \begin{small}
  \renewcommand\arraystretch{1.25}
  \resizebox{\textwidth}{!}{ 
  \begin{tabular}{l|ccc|ccc|c|c}
    \toprule
    \midrule
    \multicolumn{1}{c|}{\multirow{2}{*}{}} &
    \multicolumn{3}{c|}{\emph{In-Distribution}} & \multicolumn{3}{c|}{\emph{Cross-Size}} & \multicolumn{1}{c|}{\emph{Cross-Distribution}} & \multicolumn{1}{c}{\emph{Cross-Size \& Distribution}} \\
     & mixed\_d3\_n1000 & mixed\_d5\_n1000 & mixed\_d7\_n1000 & mixed\_d3\_n2000 & mixed\_d5\_n2000 & mixed\_d7\_n2000 & cluster\_n1000 & cluster\_n2000 \\
    \midrule
     L2D (512) & 142.53 (2) & 119.33 (2) & \textbf{102.96 (5)} & 287.64 (2) & 245.55 (3) & 201.84 (1) & 149.43 (2) & 194.70 (2) \\
     L2D (2048) & 140.90 (1) & 161.61 (2) & 208.95 (1) & 312.07 (0) & 194.84 (3) & 300.44 (0) & \textbf{81.21 (4)} & 265.26 (1) \\
     Ours (512) & \textbf{80.95 (7)} & \textbf{95.53 (6)} & 124.95 (4) & \textbf{101.28 (8)} & \textbf{139.44 (4)} & \textbf{96.03 (9)} & 91.96 \textbf{(4)} & \textbf{97.18 (7)} \\
    \midrule
    \bottomrule
  \end{tabular}}
%   \end{sc}
  % \end{small}
  \end{center}
  \vskip -0.1in
\end{table}

\section{Discussion}
\label{app:discussion}
\textbf{Training Efficiency and Scalability.} The training efficiency and scalability of the proposed framework could be analyzed from two perspectives. 1) \emph{Meta-learning algorithm:} the second-order meta-learning algorithm is computationally expensive due to the calculation of Hessian-vector products and the needs of keeping computational graphs in memory (so that we could backward through them a second time). The improved first-order approximation method could greatly improve the training efficiency while maintaining performance. 
2) \emph{Base model:} we use POMO~\cite{kwon2020pomo}, which is a popular autoregressive construction-based method, as our base model in Section \ref{exp}. It leverages the attention mechanism and augments each training instance by starting with different nodes. Therefore, the (meta-)training of POMO is computationally expensive especially for the large-scale problems. 
In addition to the autoregressive construction-based methods, non-autoregressive construction-based and improvement-based methods also receive much attention in the literature, and they have the potential to mitigate the training efficiency and scalability issue.
Typically, non-autoregressive construction-based methods learn an edge adjacency matrix (i.e., heat-map), from which the advanced post-hoc search strategies (e.g., Monte-Carlo Tree Search) construct solutions. Improvement-based methods could leverage decomposition (or divide-and-conquer) to first solve small-scale subproblems, and then obtain the feasible solution to the global problem.
In general, they are more computationally efficient or effective to solve large-scale problems (e.g., TSP10000 in~\citet{fu2021generalize,qiu2022dimes,sun2023difusco} or VRP2000 in~\citet{li2021learning}), but at the expense of much longer inference time or extra domain knowledge (e.g., advanced search strategies). As shown in Appendix \ref{app:generalizability}, we further apply our method to L2D~\cite{li2021learning}, where we find its meta-training is much more efficient than that of POMO. For example, its second-order meta-training (on VRP500-1000) only takes 9 hours on a NVIDIA V100 GPU (32GB). Therefore, the choices of meta-learning algorithms and base models may have significant effects on the training efficiency and scalability of the proposed framework.
% Two perspectives: meta-training algorithms and backbones (autoregressive vs. non-autoregressive vs. improvement)

\textbf{Meta-training on Pretrained Models.} We would like to note that meta-training from scratch is non-trivial based on our experiments. If a pretrained model exists, it is suggested to conduct meta-training on it. Intuitively, the pretrained model could be regarded as a good initialization for the meta-learning framework (i.e., both the inner-loop and outer-loop optimization), and therefore could improve the training efficiency. 

\textbf{Performance on Small-Scale VRP Instances.} In this paper, we mainly consider evaluating the generalization performance on large-scale instances. Here, we provide empirical results on small-scale instances, which are chosen from CVRPLIB (Set-P)~\cite{augerat1995approche}, with $n\le 30$. 
The results are shown in Table \ref{sup_cvrplib_xetp}. We observe that the performance on small instances may not be necessarily good if training on large sizes (i.e., CVRP50-200). The zero-shot performance is around 0.00\%-4.17\% with greedy search. However, for small-scale instances, it is quite efficient to use advanced search strategies (e.g., EAS~\cite{hottung2022efficient}), resulting in 0.00\% on all these benchmark instances.
% On the other hand, we would like to note that most studied VRPs in literature and realistic VRPs are featured by large problem sizes with more than 50 nodes. Therefore, our framework is mainly evaluated on such practical sizes in the paper.

\textbf{Sensitivity on Training Task Distribution.} The training task set is expected to contain instances with diverse distributions. As discussed in Appendix \ref{app:data_gen}, in addition to the uniform distribution, we use the gaussian mixture distribution due to its elegance and flexibility in changing distributions (via hyperparameters) into various patterns.
Moreover, we also try to use different training task distributions. For example, we only use uniform, gaussian and cluster distributions, which are commonly used in the literature, to construct the training task set. 
Although our method outperforms baselines, all methods cannot achieve decent omni-generalization performance during evaluation. This might be led by the monotonous training tasks, which could not provide the model with sufficient or diverse information, making it hard to generalize well. 

\begin{table*}[!t]
  \vskip -0.1in
  \caption{Results on CVRPLIB (Set-P)~\cite{augerat1995approche} instances.}
  \label{sup_cvrplib_xetp}
  \vskip 0.1in
  \begin{center}
  \begin{small}
%   \begin{sc}
  \renewcommand\arraystretch{1.0}
%   \resizebox{\textwidth}{!}{ 
  \begin{tabular}{ll|cccc}
    \toprule
    \midrule
    \multicolumn{2}{c|}{} & \multicolumn{2}{c}{Ours} & \multicolumn{2}{c}{Ours+EAS}\\
     Instance  &  Opt. & Obj. & Gap & Obj. & Gap \\
    \midrule
     P-n16-k8 & 450 & \textbf{450} & \textbf{0.00\%} & \textbf{450} & \textbf{0.00\%} \\
     P-n19-k2 & 212 & 219 & 3.30\% & \textbf{212} & \textbf{0.00\%} \\
     P-n20-k2 & 216 & 225 & 4.17\% & \textbf{216} & \textbf{0.00\%} \\
     P-n21-k2 & 211 & 213 & 0.95\% & \textbf{211} & \textbf{0.00\%} \\
     P-n22-k2 & 216 & 219 & 1.39\% & \textbf{216} & \textbf{0.00\%} \\
     P-n22-k8 & 603 & 610 & 1.16\% & \textbf{603} & \textbf{0.00\%} \\
     P-n23-k8 & 529 & 548 & 3.59\% & \textbf{529} & \textbf{0.00\%} \\
    \midrule
    \bottomrule
  \end{tabular}
%   \end{sc}
  \end{small}
  \end{center}
  \vskip -0.1in
\end{table*}

%%%%%%%%%%%%%%%%%%%%%%%%%%%%%%%%%%%%%%%%%%%%%%%%%%%%%%%%%%%%%%%%%%%%%%%%%%%%%%%
%%%%%%%%%%%%%%%%%%%%%%%%%%%%%%%%%%%%%%%%%%%%%%%%%%%%%%%%%%%%%%%%%%%%%%%%%%%%%%%

\end{document}